\documentclass{article}

    \PassOptionsToPackage{numbers, compress}{natbib}

    \usepackage[final]{neurips_2024}

\usepackage[utf8]{inputenc} 
\usepackage[T1]{fontenc}    
\usepackage{hyperref}       
\usepackage{url}            
\usepackage{booktabs}       
\usepackage{amsfonts}       
\usepackage{nicefrac}       
\usepackage{microtype}      
\usepackage{xcolor}         

\usepackage{float}
\usepackage{graphicx}
\usepackage{multirow}
\usepackage{amsmath}
\usepackage{xcolor} 
\usepackage{caption}
\usepackage{pifont} 
\usepackage{color, colortbl} 
\usepackage{arydshln} 
\usepackage{tabularx} 
\usepackage{footnote} 
\newcommand{\meteor}{\includegraphics[width=0.025\textwidth]{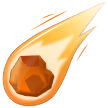}} 

\title{\includegraphics[width=0.08\textwidth]{figures/Meteor_emoji.png}\hspace{-1ex}Meteor: Mamba-based Traversal of Rationale\\for Large Language and Vision Models}

\author{
  Byung-Kwan Lee\\
  KAIST\\
  \texttt{leebk@kaist.ac.kr}
  \And
  Chae Won Kim\\
  KAIST\\
  \texttt{chaewonkim@kaist.ac.kr} \\
  \AND
  \hspace{-6.5ex} Beomchan Park\\
  \hspace{-6.5ex} KAIST\\
  \hspace{-6.5ex} \texttt{bpark0810@kaist.ac.kr} \\
  \And
  \hspace{-2.5ex}Yong Man Ro\\
  \hspace{-2.5ex} KAIST\\
  \hspace{-2.5ex} \texttt{ymro@kaist.ac.kr}
}

\begin{document}

\maketitle

\begin{abstract}
The rapid development of large language and vision models (LLVMs) has been driven by advances in visual instruction tuning. Recently, open-source LLVMs have curated high-quality visual instruction tuning datasets and utilized additional vision encoders or multiple computer vision models in order to narrow the performance gap with powerful closed-source LLVMs. These advancements are attributed to multifaceted information required for diverse capabilities, including fundamental image understanding, real-world knowledge about common-sense and non-object concepts (\textit{e.g.,} charts, diagrams, symbols, signs, and math problems), and step-by-step procedures for solving complex questions. Drawing from the multifaceted information, we present a new efficient LLVM, \textbf{M}amba-bas\textbf{e}d \textbf{t}rav\textbf{e}rsal \textbf{o}f \textbf{r}ationales (\meteor \textbf{Meteor}), which leverages multifaceted rationale to enhance understanding and answering capabilities. To embed lengthy rationales containing abundant information, we employ the Mamba architecture, capable of processing sequential data with linear time complexity. We introduce a new concept of \textit{traversal of rationale} that facilitates efficient embedding of rationale. Subsequently, the backbone multimodal language model (MLM) is trained to generate answers with the aid of rationale. Through these steps, \meteor Meteor achieves significant improvements in vision language performances across multiple evaluation benchmarks requiring diverse capabilities, without scaling up the model size or employing additional vision encoders and computer vision models. Code is available in \href{https://github.com/ByungKwanLee/Meteor}{https://github.com/ByungKwanLee/Meteor}.
\end{abstract}

\section{Introduction}
\label{sec:intro}

Following the successful zero-shot achievements of instruction-tuned large language models (LLMs)~\citep{wei2022finetuned, chung2022scaling}, visual instruction tuning~\citep{liu2023visual} has spurred the rapid development of large language and vision models (LLVMs). The emergence of closed-source LLVMs, such as GPT-4V~\citep{achiam2023gpt}, Gemini-Pro~\citep{team2023gemini}, and Qwen-VL-Plus~\citep{bai2023qwen}, has prompted several studies to create high-quality question-answer visual instruction tuning datasets~\citep{bai2023qwen, chen2023sharegpt4v, liu2023improved, dai2023instructblip, li2024mini, wang2023cogvlm} and to scale up the model sizes of open-source LLVMs~\citep{li2024mini, liu2024llavanext, young2024yi, mckinzie2024mm1, chen2023internvl}, aiming to compete with their closed-source counterparts by leveraging the scaling law~\citep{kaplan2020scaling, tay2021scale, chung2022scaling}.

Recent research trends focus on enhancing image resolution~\citep{li2023otterhd, bai2023qwen, wang2023cogvlm, ye2023mplug2, hu2024mplug} and dividing images into smaller sections~\citep{liu2024llavanext, mckinzie2024mm1, li2024mini, xu2024llava} to improve image perception capabilities. Additionally, some studies have utilized additional vision encoders~\citep{kar2024brave, lu2024deepseek, goncharova2024omnifusion, ranzinger2023radio} such as EVA-CLIP~\citep{fang2023eva}, DINOv2~\citep{oquab2023dinov2}, SAM~\citep{kirillov2023segment}, and SigLIP~\citep{zhai2023sigmoid}. Various computer vision models~\citep{chen2024spatialvlm, wang2024all, jiao2024enhancing, lee2024collavo, lee2024moai} have also been employed for tasks such as segmentation, detection, scene graph generation, and optical character recognition (OCR) to enhance LLVMs' answering capabilities with the help of external perception information.

\begin{figure}[t!]
\vspace{-5mm}
    \centering
    \includegraphics[width=\textwidth]{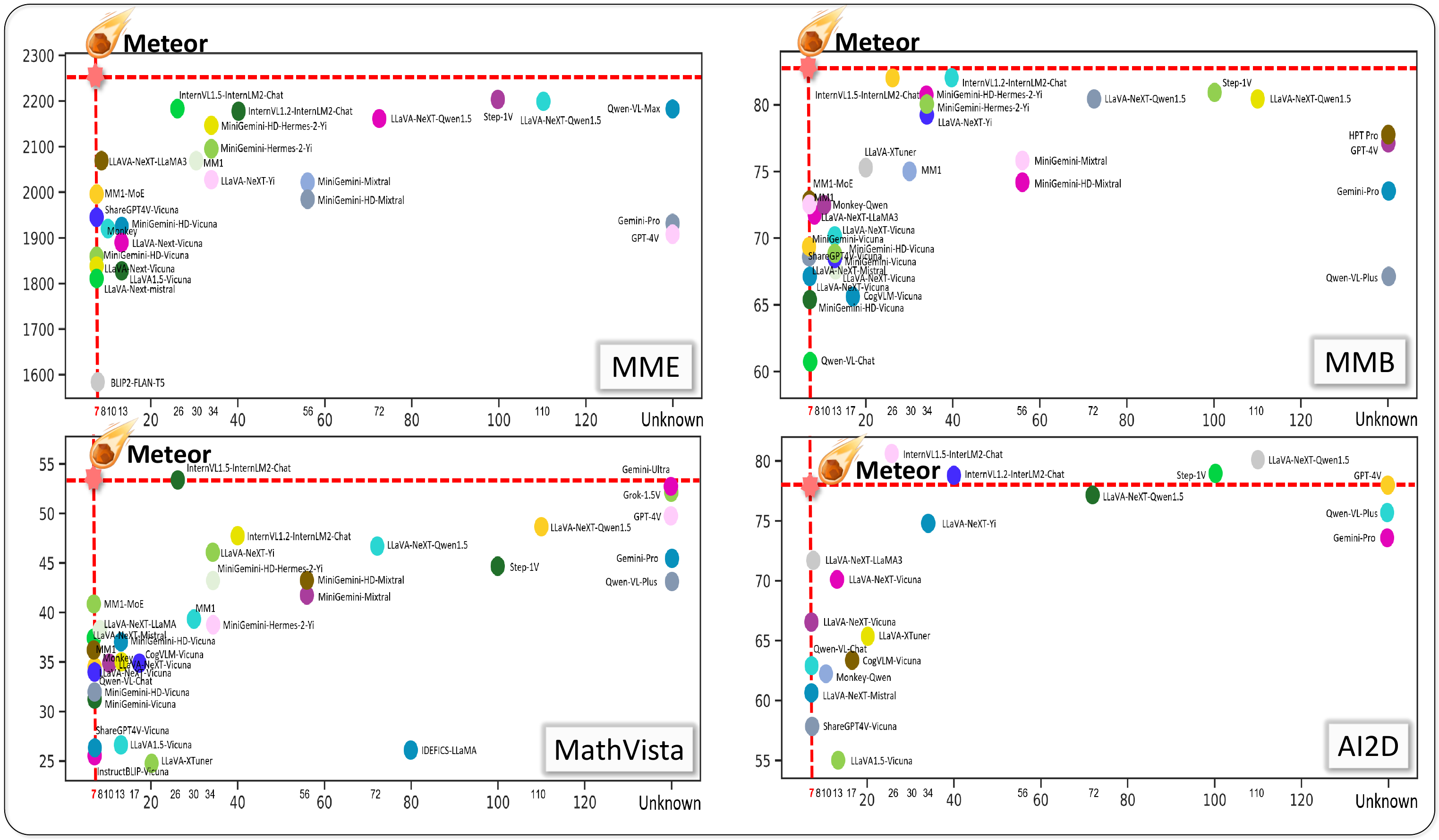}
    \caption{Across 7B to over 110B parameters, comparing lots of open- and closed-source LLVMs with \meteor Meteor on MME~\cite{fu2023mme}, MMB~\cite{liu2023mmbench}, MathVista~\cite{lu2023mathvista}, and AI2D~\cite{kembhavi2016diagram} requiring diverse capabilities for image understanding, common-sense knowledge, non-object concept understanding, etc.}
    \label{fig:figure1}
    \vspace{-5mm}
\end{figure}

These efforts, along with the curation of high-quality visual instruction datasets, have significantly reduced the performance gap between open- and closed-source LLVMs across numerous evaluation benchmarks and have even led to superior performances on some benchmarks. These successful developments are credited to the multifaceted information necessary for a wide range of capabilities. This encompasses fundamental image understanding, real-world knowledge of common-sense and non-object concepts (\textit{e.g.,} charts, diagrams, symbols, signs, and math problems), and step-by-step procedures for solving complex questions.

Inspired by the key importance of the multifaceted information, we explore the possibility of designing efficient LLVMs that implicitly embed it as a form of multifaceted rationale (See Appendix \ref{sec:appA} for more details), without significantly increasing model size and without using additional explicit vision encoders and computer vision models during the inference phase. Hence, we present a new efficient LLVM, \textbf{M}amba-bas\textbf{e}d \textbf{t}rav\textbf{e}rsal \textbf{o}f \textbf{r}ationale (\meteor \textbf{Meteor}), comprising two core components: the Mamba architecture~\citep{gu2023mamba} and a multimodal language model (MLM) based on a pretrained large language model (LLM). The multifaceted rationale has rich information for achieving diverse capabilities, so its length is inherently long. This is why we employ the Mamba architecture, hereinafter referred to as Meteor-Mamba, which takes advantage of embedding lengthy input. It serves as an embedding module for the rationale, enabling Meteor-MLM, the MLM component, to address questions with the help of these embedded rationales. When conveying the knowledge of embedded rationales from Meteor-Mamba to Meteor-MLM, we introduce a new concept of \textit{traversal of rationale}, which spurs embedding of long sequential rationales.
\begin{figure}[t!]
\vspace{-5mm}
    \centering
    \includegraphics[width=\textwidth]{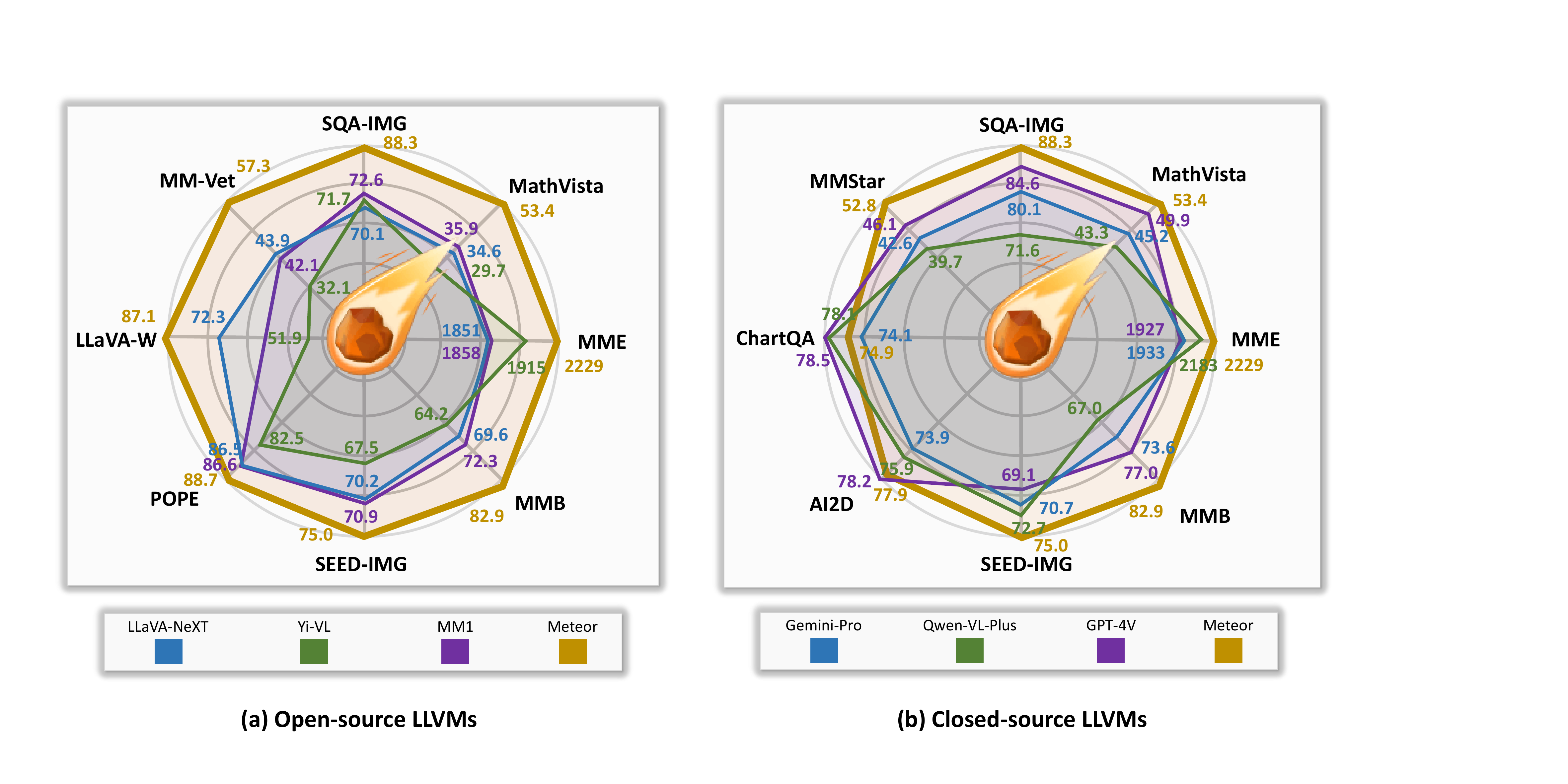}
    \vspace{-5mm}
    \caption{Overall comparison of \meteor Meteor compared with other open- and closed-source LLVMs.}
    \label{fig:figure2}
    \vspace{-5mm}
\end{figure}

To ensure that \meteor Meteor encompasses diverse capabilities for vision-language tasks (e.g., image understanding, common-sense knowledge, charts, diagrams, documents, signs, symbols, and math problems), we gather 2.1M question-answer pairs from existing visual instruction tuning datasets: ShareGPT4V-Caption/Instruct~\citep{chen2023sharegpt4v}, MiniGemini-Instruct~\citep{li2024mini}, Doc-Downstream/Reason~\citep{hu2024mplug}, GLLaVA-Align/Instruct~\citep{gao2023g}, and Math-Vision/Instruct/Plus~\citep{wang2024measuring, yue2023mammoth, yue2024mammoth2}. Subsequently, we utilize the light and fast Claude Haiku API~\citep{claude3series2024} to generate detailed and comprehensive rationales tailored for the collected 2.1M question-answer pairs. These rationales are carefully filtered by human reviewers with the aid of GPT-4V, resulting in 1.1M question-rationale-answer triples (Appendix \ref{sec:appA}).

Using the question-rationale pairs in the curated 1.1M triples, the first training step involves training Meteor-Mamba and miscellaneous projectors, \textit{i.e.,} a vision projector and tor projector. During this step, Meteor-Mamba is trained to embed long sequential rationales. In the second training step, all components of \meteor Meteor are trained using the question-answer pairs in the curated 1.1M triples. Through these steps, we demonstrate that \meteor Meteor significantly improves vision-language performance, as shown in Figure~\ref{fig:figure1}, compared with other open- and closed-source LLVMs, on numerous benchmarks requiring diverse capabilities. As illustrated in Figure~\ref{fig:figure2}, these results advocate for the possibility of building efficient LLVMs with a multifaceted rationale, beyond the scope of scaling model size, additional vision encoders, and multiple computer vision models.

Our contribution can be summarized into two main aspects:
\begin{itemize}
\item We introduce a new efficient large language and vision model (LLVM), \textbf{M}amba-bas\textbf{e}d \textbf{t}rav\textbf{e}rsal \textbf{o}f \textbf{r}ationale (\meteor \textbf{Meteor}), which comprehends long sequential rationales under a new concept of \textit{traversal of rationale} and predict answers with the help of the rationale.
\item Despite its efficient model size, \meteor \textbf{Meteor} showcases significant advancements across various evaluation benchmarks requiring diverse capabilities for image understanding, common-sense, non-object concepts, and more.
\end{itemize}

\section{Related Works}
\label{sec:related}

\paragraph{Rationale-Guided Prediction.} Behind the answers of large language models (LLMs), rationale has played a significant role in enhancing answering capabilities across a myriad of natural language processing and vision language tasks in various forms such as (a) human annotation \citep{strout2019human, lu2022rationale}, (b) knowledge distillation \citep{hsieh2023distilling, wang2023scott, xiong2023rationale, li2023symbolic, kang2024knowledge}, and (c) chain-of-thought (CoT) \citep{wei2022chain, zhang2022automatic, shum2023automatic, krishna2024post}. Rationale mimics the human thought process, providing explanations or justifications before answering questions. (a) Strout \textit{et al.}\citep{strout2019human} and Lu \textit{et al.}\citep{lu2022rationale} integrate human-annotated rationale with LLMs to fortify their performance, thereby enhancing model robustness against out-of-distribution scenarios with human annotations. (b) In knowledge distillation, rationale is used to effectively distill LLMs \citep{hsieh2023distilling, wang2023scott, xiong2023rationale, li2023symbolic, kang2024knowledge} into smaller language models. They first extract rationale from LLMs and then fine-tune smaller language models with the extracted rationale, demonstrating their efficacy through various evaluation benchmarks. (c) To directly apply rationale to LLMs, researchers have used a `think step-by-step' prompt called Chain of Thought (CoT) \citep{wei2022chain}. When combined with few-shot learning, CoT elicits step-by-step rationale directly from LLMs for an input question followed by a series of questions, rationales, and answers used as few-shot examples. This has been further streamlined by automating rationale generation in CoT prompts, as explored by \citep{zhang2022automatic} and \citep{shum2023automatic}, eliminating the need for human-annotated rationale for few-shot examples.

Similar to (b), we also leverage the power of closed-source LLVMs but, in contrast, employ another model, Meteor-Mamba, to embed the rationale, instead of directly training Meteor-MLM to generate the created rationales. In other words, we separate the roles of the models where Meteor-MLM generates the answer, while Meteor-Mamba embeds the rationale, which encompasses diverse capabilities: fundamental image understanding, incorporation of real-world knowledge of common-sense, understanding of non-object concepts (e.g., charts, diagrams, symbols, signs, and math), and following systematic step-by-step procedures for solving complex questions. Once accompanied by LLVMs, their answer capabilities are expected to improve with the help of the rationale.

\paragraph{Large Language and Vision Models.} Following the emergence of visual instruction tuning datasets created by LLaVA~\cite{liu2023visual, liu2023improved, liu2024llavanext} and InstructBLIP~\cite{dai2023instructblip}, there has been rapid development of large language and vision models (LLVMs): Shikra~\cite{chen2023shikra}, IDEFICS~\cite{laurenccon2023obelisc}, Qwen-VL~\cite{bai2023qwen}, MiniGPT-4~\cite{zhu2023minigpt}, Otter~\cite{li2023otter}, mPLUG-Owl~\cite{ye2023mplug, ye2023mplug2}, ShareGPT4V~\cite{chen2023sharegpt4v}, LLaVA-XTuner~\cite{2023xtuner}, Intern-XC~\cite{zhang2023internlm}, MM1~\cite{mckinzie2024mm1}, MiniGemini~\cite{li2024mini}, InternVL Families~\cite{chen2023internvl, chen2024far}, along with efforts to gather or curate high-quality visual instruction tuning datasets for various purposes: ShareGPT4V~\cite{chen2023sharegpt4v}, ALLaVA~\cite{chen2024allava}, MiniGemini~\cite{li2024mini}, mPLUG-DocOwl~\cite{hu2024mplug}, GLLaVA~\cite{gao2023g}, MathVision~\cite{wang2024measuring}, MathInstruct~\cite{yue2023mammoth}, and MathPlus~\cite{yue2024mammoth2}. Recently, Otter-HD~\cite{li2023otterhd}, Qwen-VL~\cite{bai2023qwen}, CogVLM~\cite{wang2023cogvlm}, and mPLUG Families~\cite{ye2023mplug2, hu2024mplug} have increased image resolution. Additionally, LLaVA-NeXT~\citep{liu2024llavanext}, MM1~\cite{mckinzie2024mm1}, and MiniGemini~\cite{li2024mini} divide images into smaller sections, while LLaVA-UHD~\cite{xu2024llava} and InternVL1.5~\cite{chen2024far} employ dynamically split image partitions depending on their sizes. These research trends aim to improve image perception capabilities, thereby enhancing understanding of images and natural language instructions. Furthermore, BRAVE~\cite{kar2024brave}, DeepSeek-VL~\cite{lu2024deepseek}, OmniFusion~\cite{goncharova2024omnifusion}, MoVA~\cite{zong2024mova}, and AM-RADIO~\cite{ranzinger2023radio} have utilized additional vision encoders such as EVA-CLIP~\citep{fang2023eva}, DINOv2~\citep{oquab2023dinov2}, SAM~\citep{kirillov2023segment}, and SigLIP~\citep{zhai2023sigmoid}. Apart from those, SpatialVLM~\cite{chen2024spatialvlm}, ASMv2~\cite{wang2024all}, LAR/LAF~\cite{jiao2024enhancing}, CoLLaVO~\cite{lee2024collavo}, and MoAI~\cite{lee2024moai} employ multiple computer vision models for tasks such as segmentation, detection, scene graph generation, and optical character recognition (OCR).

We view this series of efforts as procedures for enlarging LLVM's knowledge space regarding multifaceted information. From a fresh perspective, we believe that embedding it in the form of a multifaceted rationale serves as a key in developing efficient LLVMs, where they include in-depth explanations required for acquiring diverse capabilities. From this perspective, \meteor Meteor is expected to inherently embed the multifaceted rationale and improve answering capabilities with the help of the embedded rationale, even without significantly increasing model size or relying on additional explicit vision encoders and computer vision models.

\section{\includegraphics[width=0.06\textwidth]{figures/Meteor_emoji.png}\hspace{-1ex}Meteor: Mamba-based traversal of rationale}
\label{sec:meteor}

\begin{figure}[t!]
\vspace{-8mm}
    \centering
    \includegraphics[width=\textwidth]{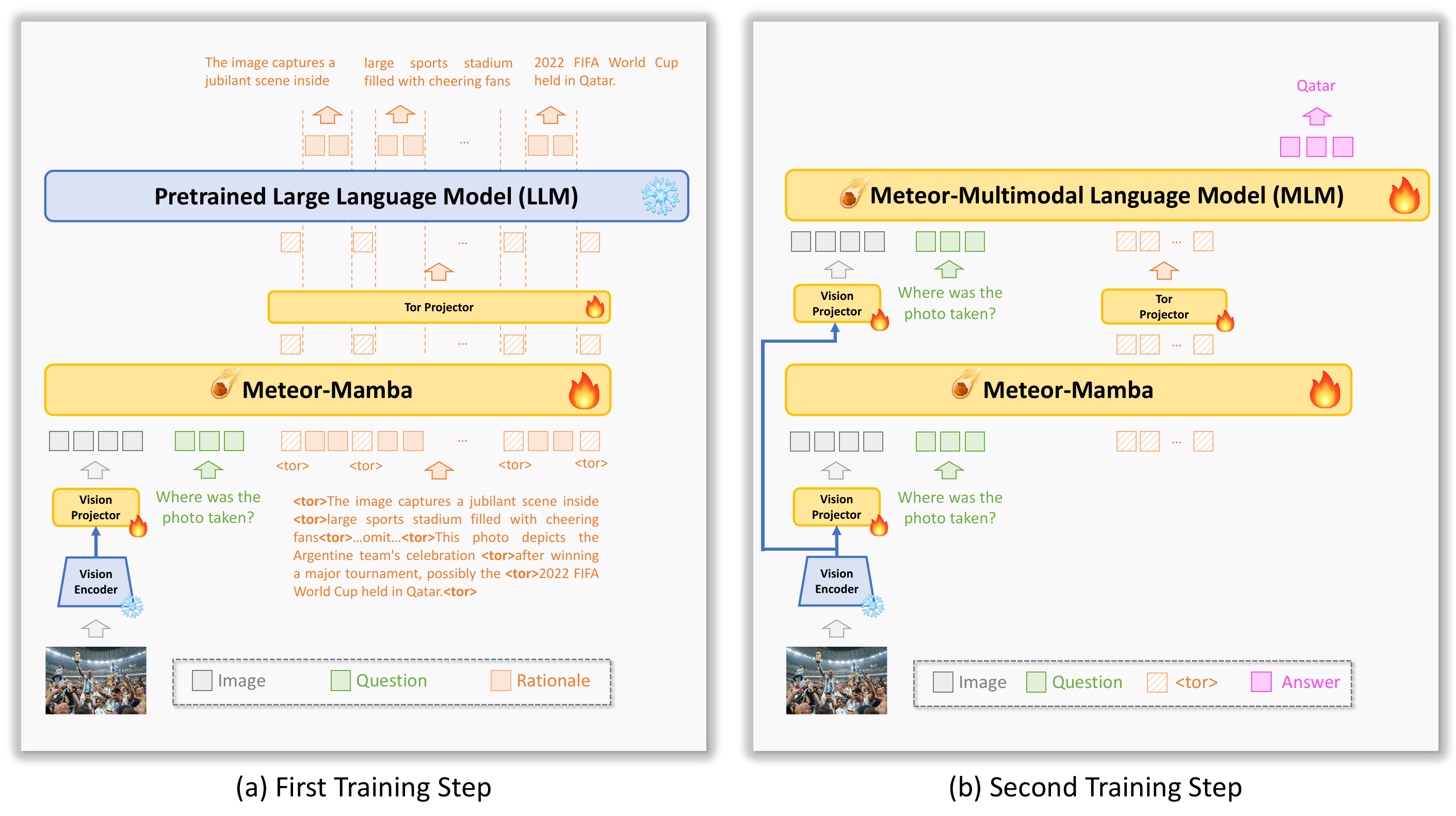}
    \caption{Overview of \meteor Meteor architecture and its training steps. Note that, `Meteor-Multimodal Language Model (MLM)' indicates that as training progresses, the pretrained language model evolves into a multimodal language model.}
    \label{fig:figure3}
    \vspace{-5mm}
\end{figure}

\paragraph{Model Architecture.} As illustrated in Figure~\ref{fig:figure3}, \meteor Meteor comprises a vision encoder, vision projector, Mamba architecture~\cite{gu2023mamba}, tor projector, and backbone multimodal language model (MLM) based on a pretrained large language model (LLM). For the vision encoder, we use CLIP-L/14~\cite{clip}, which is a text-aligned vision module that takes advantage of image understanding adeptness powered by text descriptions. For the vision projector and tor projector, we employ MLP modules containing two fully-connected layers with the GELU activation function~\cite{hendrycks2016gaussian}. Next, we use the Mamba-130M architecture for computational efficiency and adopt InternLM2-7B~\cite{2023internlm, cai2024internlm2}, learned with 2T tokens of multilingual text data in RLHF~\cite{ouyang2022training}, as the backbone large language model (LLM).

\paragraph{Configuration of Visual Instruction Tuning Dataset.} To build a visual instruction tuning set, we cover not only fundamental image understanding but also a wide range of diverse capabilities: common-sense knowledge, non-object concepts (\textit{e.g.,} charts, diagrams, documents, signs, symbols, and math problems), cognitive reasoning, multi-discipline tasks, and integrated abilities. For the question-answer visual instruction tuning dataset, we choose 664K question-answer pairs in ShareGPT4V-Instruct~\cite{chen2023sharegpt4v}, including LLaVA-Instruct-665K~\cite{liu2023improved}. Additionally, in ShareGPT4V-Caption~\cite{chen2023sharegpt4v}, we select 91K image descriptions for images from LAION~\cite{schuhmann2022laion}, CC~\cite{changpinyo2021conceptual}, SBU~\cite{saleh2015large}, MS-COCO~\cite{lin2014microsoft}, TextCaps~\cite{sidorov2020textcaps}, and web images~\cite{ordonez2011im2text, schuhmann2021laion, sharma2018conceptual} that depict landmarks, animals, celebrities, art, text, and nature. The selected question-answer pairs primarily focus on fundamental image understanding and common-sense knowledge, with fewer data samples covering non-object concepts, cognitive reasoning, multi-discipline tasks, and integrated abilities. To strengthen these areas, we selectively gather 27K question-answer pairs of DocVQA~\cite{mathew2021docvqa}, ChartQA~\cite{masry2022chartqa}, DVQA~\cite{kafle2018dvqa}, and AI2D~\cite{kembhavi2016diagram} from MiniGemini-Instruct~\cite{li2024mini}. Moreover, we use 574K/27K question-answer pairs of DeepForm~\cite{svetlichnaya2020deepform}, InfoVQA~\cite{mathew2022infographicvqa}, DocVQA~\cite{mathew2021docvqa}, KleisterCharity~\cite{stanislawek2021kleister}, TabFact~\cite{chen2019tabfact}, TextVQA~\cite{singh2019towards}, WikiTable~\cite{pasupat-liang-2015-compositional}, TextCaps~\cite{sidorov2020textcaps}, and VisualMRC~\cite{tanaka2021visualmrc} from Doc-Downstream/Reason~\cite{hu2024mplug}. To achieve broad coverage of math knowledge, we also include 177K GLLaVA-Align/Instruct~\cite{gao2023g}, 3K MathVision~\cite{wang2024measuring}, and 566K text-only samples from Math-Instruct/Plus~\cite{yue2023mammoth, yue2024mammoth2}. In summary, we collect 755K real-world images, 627K images for documents, charts, diagrams, signs, and symbols, and 747K math samples (180.5K with images and 566.8K text-only). Overall, the question-answer visual instruction tuning samples sum up to 2.1M.

\paragraph{Curating Rationale.} Using the gathered 2.1M question-answer pairs, we utilize the light and fast Claude Haiku API~\cite{claude3series2024} to generate detailed and comprehensive rationales. We use the prompt template: "\textit{Question: }\{\}. \textit{Answer: }\{\}. \textit{Based on the question and answer, carefully provide an explanation about how to answer the question in detail.}" Here, \{\} represents the placeholder for the corresponding language description. Afterward, we assess the rationale score from GPT-4V~\cite{achiam2023gpt} using the following template: "\textit{Question: }\{\}. \textit{Rationale: }\{\}. \textit{Answer: }\{\}. \textit{Based on the question, rationale, and answer, provide a score from 0 to 10, evaluating how well the rationale is described to solve the question. If the given rationale is insufficient, you should rigorously give a score below 5.}" Subsequently, we filter out the generated rationales with a score below 5. The rationales that pass this automated evaluation are then subjected to human review to determine \textit{{Yes or No}} on whether they provide a proper description to address the question. Finally, this series of processes yields 1.1M question-rationale-answer triples, which include 338K real-world images covering common-sense knowledge and a few samples for diverse capabilities, 379K images for documents, charts, diagrams, signs, and symbols, and 342K math samples (165K with images and 177K text-only).

\paragraph{Mamba Architecture.}
\label{para:mamba arch}
To make LLVMs inherently possess rationale when addressing complex questions, we generate comprehensive rationales based on question-answer pairs. Subsequently, we employ the Mamba architecture~\citep{gu2023mamba}, leveraging its capability to handle lengthy rationales while maintaining computational efficiency. This approach allows us to effectively incorporate the rationale in an environment where the curated 1.1M question-rationale pairs have an average length of 213 tokens, which is approximately ten times longer than the average length of 22 tokens of ground truth answers in typical visual instruction tuning datasets~\cite{xu2024llava, chen2023sharegpt4v}.

\paragraph{Traversal of Rationale.} However, it is crucial to note that we cannot acquire and utilize rationales during the inference phase without API-based models, since only user questions are given. Therefore, we propose a new concept called \textit{traversal of rationale} to effectively provide the rationale to Meteor-MLM without any help from external APIs in the inference phase. Inspired by retrieval-based knowledge~\cite{yasunaga2022retrieval}, we introduce a special token, <tor> (stands for \textbf{t}raversal \textbf{o}f \textbf{r}ationale), and evenly distribute 10 fixed number of <tor> tokens, as described in Figure~\ref{fig:figure3}. The rationale planted with <tor> is propagated into Meteor-Mamba along with image and question tokens, and then the output features in Meteor-Mamba are directly propagated into Meteor-MLM. Here, we autoregressively train Meteor to generate the part of the rationale between <tor>, whenever Meteor sees the special token <tor>. This procedure ensures that each <tor> represents the following rationale part until the next <tor> is encountered. Using a single <tor> token to encompass the rationale may not work well when embedding lengthy rationales, and if we do not consider distributing <tor> tokens in the rationale, then the later token does not refer to the earlier ones well due to the common problem of the autoregressive mechanism's forgetting nature~\cite{gu2023mamba}. This is why we place multiple <tor> tokens in the rationale instead of one.

\paragraph{Training Strategy.} In the first training step, we leverage the question-rationale pairs in the curated 1.1M triples to train Meteor-Mamba and miscellaneous projectors. Throughout this step, the long sequential rationale is embedded into Meteor-Mamba through traversal of rationale by autoregressively generating rationale parts between the special tokens <tor>. By freezing Meteor-MLM, Meteor-Mamba seamlessly incorporates the rationale. In the second training step, we utilize the question-answer pairs in the curated 1.1M triples to jointly train Meteor-Mamba, Meteor-MLM, and the miscellaneous projectors. Here, multiple <tor> special tokens are only propagated to Meteor-Mamba. Then, the rationale-embedded features in Meteor-Mamba corresponding to the special tokens <tor> are only fed into Meteor-MLM, enabling it to adeptly answer complex questions, even in the absence of explicit rationale descriptions. In essence, these steps equip \meteor Meteor with the capability to effectively address complex questions with the aid of the rationale.

\definecolor{Gray}{gray}{0.93}
\definecolor{Green}{rgb}{0.9, 0.95, 0.97}
\newcommand{\cmark}{\ding{51}}%
\newcommand{\xmark}{\ding{55}}%

\begin{table}[t!]
\vspace{-8mm}
\centering
\resizebox{\linewidth}{!}{
\renewcommand{\tabcolsep}{0.8mm}
\begin{tabular}{lccccccccccccc}
\toprule
LLVMs     & Q-Bench & SQA$^{\text{I}}$ & AI2D & ChartQA & SEED$^{\text{I}}$ & POPE & HallB & MME & MathVista & MMB & MMB$^{\text{CN}}$ & MM-Vet & LLaVA$^{\text{W}}$  \\
\midrule
BLIP2-13B~\cite{blip2}                            & -    & 61.0 & -    & -    & 46.4 & 85.3 & -    & 1584 & -    & -    & -    & 22.4 & -    \\
InstructBLIP-7B~\cite{dai2023instructblip}        & 56.7 & 60.5 & -    & -    & 53.4 & -    & 53.6 & -    & 25.3 & 36.0 & 23.9 & 26.2 & -    \\
InstructBLIP-13B~\cite{dai2023instructblip}       & -    & 63.1 & -    & -    & -    & 78.9 & -    & -    & -    & 33.9 & -    & 25.6 & -    \\
IDEFICS-9B~\cite{laurenccon2023obelisc}           & 51.5 & -    & -    & -    & -    & 74.6 & -    & 1353 & 19.8 & 48.2 & 25.2 & 23.7 & -    \\
Qwen-VL-7B~\cite{bai2023qwen}                     & 59.4 & 67.1 & -    & -    & -    & -    & -    & -    & -    & 38.2 & 7.4  & -    & -    \\
Qwen-VL-Chat-7B~\cite{bai2023qwen}                & 33.8 & 68.2 & -    & -    & 58.2 & -    & 56.4 & 1849 & -    & 60.6 & 56.7 & 47.3 & -    \\
MiniGPT-4-7B~\cite{zhu2023minigpt}                & 51.8 & -    & -    & -    & -    & -    & -    & -    & 23.1 & 23.0 & 11.9 & 22.1 & -    \\
Otter-7B~\cite{li2023otter}                       & 47.2 & -    & -    & -    & -    & 72.5 & -    & 1599 & 19.7 & 48.3 & -    & 24.7 & -    \\
\rowcolor{Gray}
LLaVA-7B~\cite{liu2023visual}                     & -    & 38.5 & -    & -    & -    & 80.2 & 44.1 & 1055 & -    & 34.1 & 14.1 & 26.7 & -    \\
\rowcolor{Gray}
LLaVA1.5-7B~\cite{liu2023improved}                & 60.1 & 66.8 & -    & -    & 58.6 & 85.9 & -    & 1805 & -    & 64.3 & 58.3 & 30.5 & 63.4 \\
\rowcolor{Gray}
LLaVA1.5-13B~\cite{liu2023improved}               & 61.4 & 71.6 & 54.8 & 18.2 & 61.6 & 85.9 & 46.7 & 1826 & 27.6 & 67.7 & 63.6 & 35.4 & -    \\
mPLUG-Owl-7B~\cite{ye2023mplug}                   & 58.9 & -    & -    & -    & -    & -    & -    & -    & 22.2 & 46.6 & -    & -    & -    \\
mPLUG-Owl2-7B~\cite{ye2023mplug2}                 & 62.9 & 68.7 & -    & -    & -    &      & -    & -    & -    & 64.5 & 60.3 & 36.2 & -    \\
ShareGPT4V-7B~\cite{chen2023sharegpt4v}           & 63.4 & 68.4 & -    & -    & 69.7 & -    & 49.8 & 1944 & 25.8 & 68.8 & 62.2 & 37.6 & -    \\
InternLM-XC-7B~\cite{zhang2023internlm}           & 64.4 & -    & -    & -    & 66.1 & -    & 57.0 & 1919 & 29.5 & 74.4 & 72.4 & 35.2 & -    \\
Monkey-10B~\cite{li2023monkey}                    & -    & 69.4 & -    & -    & 68.9 & -    & 58.4 & 1924 & 34.8 & 72.4 & 67.5 & 33.0 & -    \\
VILA-7B~\cite{lin2023vila}                        & -    & 68.2 & -    & -    & 61.1 & 85.5 & -    & -    & -    & 68.9 & 61.7 & 34.9 & -    \\
VILA-13B~\cite{lin2023vila}                       & -    & 73.7 & -    & -    & 62.8 & 84.2 & -    & -    & -    & 70.3 & 64.3 & 38.8 & -    \\
SPHINX-7B~\cite{lin2023sphinx}                    & -    & 70.6 & -    & -    & 71.6 & 86.9 & -    & 1797 & 27.8 & 65.9 & 57.9 & 40.2 & -    \\
SPHINX-MoE-7B$\times$8~\cite{gao2024sphinx}       & 66.2 & 70.6 & -    & -    & 73.0 & \textbf{89.6} & -    & 1852 & 42.7 & 71.3 & -    & 40.9 & -    \\
SPHINX-Plus-13B~\cite{gao2024sphinx}              & 66.2 & 70.6 & -    & -    & 74.8 & 89.1 & 52.1 & 1741 & 36.8 & 71.0 & -    & 47.9 & -    \\
\rowcolor{Gray}
LLaVA-NeXT-7B~\cite{liu2024llavanext}             & -    & 70.1 & -    & -    & 70.2 & 86.5 & -    & 1851 & 34.6 & 69.6 & 63.3 & 43.9 & 72.3 \\
\rowcolor{Gray}
LLaVA-NeXT-8B~\cite{liu2024llavanext}             & -    & -    & 71.6 & 69.5 & -    & -    & -    & 1972 & 37.5 & 72.1 & -    & -    & 80.1 \\
\rowcolor{Gray}
LLaVA-NeXT-13B~\cite{liu2024llavanext}            & -    & 73.6 & 70.0 & 62.2 & 72.2 & 86.7 & -    & 1892 & 35.1 & 70.0 & 68.5 & 47.3 & 72.3 \\
MM1-7B~\cite{mckinzie2024mm1}                     & -    & 72.6 & -    & -    & 69.9 & 86.6 & -    & 1858 & 35.9 & 72.3 & -    & 42.1 & -    \\
MM1-MoE-7B$\times$32~\cite{mckinzie2024mm1}       & -    & 74.4 & -    & -    & 70.9 & 87.8 & -    & 1992 & 40.9 & 72.7 & -    & 45.2 & -    \\
MiniGemini-HD-7B~\cite{li2024mini}                & -    & -    & -    & -    & -    & -    & -    & 1865 & 32.2 & 65.8 & -    & 41.3 & -    \\
MiniGemini-HD-13B~\cite{li2024mini}               & -    & -    & -    & -    & -    & -    & -    & 1917 & 37.0 & 68.6 & -    & 50.5 & -    \\
\midrule
\rowcolor{Green}
Meteor-7B   & \textbf{69.2} 
            & \textbf{88.3}
            & \textbf{77.9}    
            & \textbf{74.9}
            & \textbf{75.0} 
            & 88.7
            & \textbf{60.4}    
            & \textbf{2229}
            & \textbf{53.4}
            & \textbf{82.9} 
            & \textbf{82.1}  
            & \textbf{57.3} 
            & \textbf{87.1}    \\
\bottomrule 
\end{tabular}
}
\vspace{2mm}
\caption{Comparison with the current existing open-source LLVMs, evaluating vision language performances of \meteor Meteor on numerous evaluation benchmarks requiring diverse capabilities: Q-Bench~\cite{wu2023q}, SQA$^{\text{I}}$~\cite{lu2022learn}, AI2D~\cite{kembhavi2016diagram}, ChartQA~\cite{masry2022chartqa}, SEED$^{\text{I}}$~\cite{li2023seed}, POPE~\cite{li2023evaluating}, HallB~\cite{liu2023hallusionbench}, MME~\cite{fu2023mme}, MathVista~\cite{lu2023mathvista}, MMB~\cite{liu2023mmbench}, MMB$^{\text{CN}}$~\cite{liu2023mmbench}, MM-Vet~\cite{yu2023mm}, and LLaVA$^{\text{W}}$~\cite{liu2023visual}. Note that, AI2D and ChartQA performances for LLaVA family models are evaluated under zero-shot conditions, while \meteor Meteor uses training dataset for them.}
\vspace{-5mm}
\label{tab:zeroshot}
\end{table}

\begin{table}[t!]
\vspace{-8mm}
\centering
\begin{minipage}[t]{\linewidth}

{\begin{minipage}[t]{0.49\linewidth}
\resizebox{\linewidth}{!}{
\renewcommand{\tabcolsep}{1mm}
\begin{tabular}{lccccccc}
\toprule
LLVMs            & CP                       & FP                       & IR                       & LR                       & ST                       & MA                       & Avg                      \\
\midrule
Yi-VL-34B~\cite{young2024yi}        & \multicolumn{1}{l}{53.2} & \multicolumn{1}{l}{31.2} & \multicolumn{1}{l}{52.0} & \multicolumn{1}{l}{32.4} & \multicolumn{1}{l}{12.4} & \multicolumn{1}{l}{35.2} & \multicolumn{1}{l}{36.1} \\
\cdashline{1-8}\noalign{\vskip 0.5ex}
CogVLM-Chat-17B~\cite{wang2023cogvlm}  & \multicolumn{1}{l}{66.8} & \multicolumn{1}{l}{36.8} & \multicolumn{1}{l}{49.2} & \multicolumn{1}{l}{31.2} & \multicolumn{1}{l}{23.6} & \multicolumn{1}{l}{11.6} & \multicolumn{1}{l}{36.5} \\
\cdashline{1-8}\noalign{\vskip 0.5ex}
SPHINX-MoE-7B$\times$8~\cite{gao2024sphinx} & 58.4                     & 40.8                     & 47.6                     & 35.2                     & 19.2                     & 32.0                     & 38.9                     \\
\cdashline{1-8}\noalign{\vskip 0.5ex}
InternVL1.2-40B~\cite{chen2023internvl}  & 67.6                     & 43.2                     & 61.2                     & 47.2                     & 24.0                     & 19.2                     & 43.7                     \\
\cdashline{1-8}\noalign{\vskip 0.5ex}
LLaVA-NeXT-34B~\cite{liu2024llavanext}   & 66.4                     & \textbf{52.0}            & 62.4                     & 46.0                     & 32.4                     & \textbf{53.6}            & 52.1                     \\
\midrule
\rowcolor{Green}
Meteor-7B        & \textbf{69.6}            & 45.6                     & \textbf{63.6}            & \textbf{53.2}            & \textbf{42.0}            & 42.8                     & \textbf{52.8}        \\
\bottomrule
\end{tabular}
}
\vspace{2mm}
\caption*{(a) MMStar~\cite{chen2024we}}
\end{minipage}
\begin{minipage}[t]{0.49\linewidth}
\resizebox{\linewidth}{!}{
\renewcommand{\tabcolsep}{1mm}
\begin{tabular}{lccccccc}
\toprule
LLVMs          & TD   & TL   & TO   & VI   & VD   & VO   & Avg  \\
\midrule
G-LLaVA-7B~\cite{gao2023g}                 & 20.9 & 20.7 & 21.1 & 17.2 & 16.4 & 9.4  & 16.6 \\
\cdashline{1-8}\noalign{\vskip 0.5ex}
LLaVA-NeXT-13B~\cite{liu2024llavanext}     & 12.8 & 12.0 & 9.9  & 10.7 & 9.7  & 6.3  & 10.3 \\
\cdashline{1-8}\noalign{\vskip 0.5ex}
ShareGPT4V-13B~\cite{chen2023sharegpt4v}   & 16.2 & 16.2 & 6.6  & 15.5 & 13.8 & 3.7  & 13.1 \\
\cdashline{1-8}\noalign{\vskip 0.5ex}
SPHINX-Plus-13B~\cite{gao2024sphinx}        & 13.9 & 11.6 & 14.9 & 11.6 & 13.5 & 10.4 & 12.2 \\
\cdashline{1-8}\noalign{\vskip 0.5ex}
SPHINX-MoE-7B$\times$8~\cite{gao2024sphinx}        & \textbf{26.2} & 17.4 & 26.7 & 16.7 & 12.5 & 11.1 & 16.8 \\
\midrule
\rowcolor{Green}
Meteor-7B                                  & 25.5 & \textbf{21.7} & \textbf{27.4} & \textbf{21.7} & \textbf{19.2} & \textbf{14.7} & \textbf{21.7} \\
\bottomrule
\end{tabular}
}
\vspace{2mm}
\caption*{(b) MathVerse~\cite{zhang2024mathverse}}
\end{minipage}
}

\begin{minipage}[t]{0.99\linewidth}
\newcolumntype{g}{>{\columncolor{Green}}c}
\resizebox{\linewidth}{!}{
\renewcommand{\tabcolsep}{1mm}
\begin{tabular}{lcccccccg}
\toprule
Benchmarks & OmniFusion~\cite{goncharova2024omnifusion} & DeepSeek-VL~\cite{lu2024deepseek} & MoVA~\cite{kar2024brave} & ASMv2~\cite{wang2024all} & LAF~\cite{jiao2024enhancing} & CoLLaVO~\cite{lee2024collavo} & MoAI~\cite{lee2024moai} & {   Meteor   } \\
\midrule
POPE       & 87.2      & 88.1  & 88.6  & 86.3  & \textbf{88.8}   & 87.2    & 87.1 & 88.7  \\
\cdashline{1-9}\noalign{\vskip 0.5ex}
SQA-IMG    & 69.2      & 57.7  & 74.4  & 87.1  & -      & 80.7    & 83.5 & \textbf{88.3}  \\
\cdashline{1-9}\noalign{\vskip 0.5ex}
LLaVA-W    & -         & -     & -     & 78.9  & -      & 69.5    & 71.9 & \textbf{87.1}  \\
\cdashline{1-9}\noalign{\vskip 0.5ex}
MM-Vet     & 39.4      & 41.5  & -     & 41.3  & 38.9   & 40.3    & 43.7 & \textbf{57.3}  \\
\cdashline{1-9}\noalign{\vskip 0.5ex}
MMStar     & -         &  -    & -     & -     & -      & 42.1    & 48.7 & \textbf{52.8}  \\
\bottomrule
\end{tabular}
}
\vspace{2mm}
\caption*{(c) Comparison with LLVMs using additional vision encoders and computer vision models}
\end{minipage}

\begin{minipage}[t]{0.99\linewidth}
\resizebox{\linewidth}{!}{
\renewcommand{\tabcolsep}{2mm}
\begin{tabular}{lccccccc}
\toprule
LLVMs      & Recognition & OCR  & Knowledge & Language Generation & Spatial Awareness & Math Problems & Avg \\
\midrule
CoLLaVO-7B~\cite{lee2024collavo} & 45.6        & 31.1 & 29.8      & 30.2                & 37.9              & 5.8  & 41.0 \\
\cdashline{1-8}\noalign{\vskip 0.5ex}
MoAI-7B~\cite{lee2024moai}    & 48.3        & 34.8 & 33.5      & 33.0                & 39.7              & 7.7  & 43.7 \\
\midrule
\rowcolor{Green}
Meteor-7B w.o. Meteor-Mamba	  & 44.5        & 33.5 & 41.8      & 31.3                & 38.6              & 29.2 & 44.8 \\
\rowcolor{Green}
Meteor-7B  & \textbf{54.1}        & \textbf{60.1} & \textbf{44.2}      & \textbf{45.0}                & \textbf{59.3}              & \textbf{57.7} & \textbf{57.3} \\
\bottomrule
\end{tabular}
}
\vspace{2mm}
\caption*{(d) Evaluating sub-benchmarks in MM-Vet~\cite{yu2023mm} with LLVMs utilizing computer vision models}
\end{minipage}

\end{minipage}
\caption{Detailed comparison of \meteor Meteor across more challenging evaluation benchmarks.}
\label{tab:star}
\vspace{-5mm}
\end{table}

\section{Experiment}
\label{sec:experi}

\paragraph{Implementation Detail.} To ensure successful reproducibility, we outline three crucial technical details of \meteor Meteor: the structure of (a) Meteor-Mamba and Meteor-MLM, (b) vision encoder and miscellaneous projectors, and (c) training and inference details.

\textbf{(a)} To build Meteor-Mamba, we use the Mamba architecture with 24 layers and a 768 hidden dimension, resulting in a total of 130M parameters, which is relatively trivial compared to the approximately 7B parameters of the pretrained InternLM2-7B~\cite{2023internlm, cai2024internlm2}. It is executed under the efficient computation of hardware-aware state expansion~\cite{gu2023mamba}, where we borrow the tokenizer~\cite{kudo2018sentencepiece} from the backbone MLM to fit the language expression space in the backbone MLM. Meteor-MLM is based on InternLM2-7B~\cite{2023internlm, cai2024internlm2} with 32 layers and a 4096 hidden dimension.

\textbf{(b)} We use a vision encoder with 428M CLIP-L/14~\cite{clip}, which has 24 layers and a 1024 hidden dimension. The resolution of the positional embedding is interpolated from $24\times 24$ to $35\times 35$ to accommodate a $490\times 490$ image resolution. The vision projector involves an MLP that adapts the hidden dimension from 1024 to 4096 to fit that of the backbone MLM. Similarly, we build the tor projector to convey embedded rationales from Meteor-Mamba into Meteor-MLM, employing the same structure as the vision projector but transferring the hidden dimension from 768 to 4096.

\textbf{(c)} We train and evaluate \meteor Meteor in the following computing environment: Intel(R) Xeon(R) Gold 6230, 256 GB RAM, and 8$\times$NVIDIA RTX A6000 48GB VRAM. To efficiently train it, we use one epoch of training for each training step under 4-bit quantization and bfloat16 data type~\cite{kalamkar2019study} for Meteor-MLM, where double quantization and normalized float 4-bit (nf4)~\cite{dettmers2023qlora} are used. Meteor-Mamba uses float32 data type because training it with bfloat16 or float16 has been reported to produce an unstable learning process. In addition, QLoRA~\cite{hu2021lora, dettmers2023qlora}  is used to train Meteor-MLM, with 64 rank and 64 alpha parameters. We use the AdamW~\cite{loshchilov2018decoupled} optimizer and schedule the learning rate by cosine annealing~\cite{loshchilov2016sgdr} from 1e-4 to 1e-6 in each training step, with gradient checkpointing~\cite{sohoni2019low} applied to Meteor-MLM for efficient memory management. With a gradient accumulation of 6, we set batch sizes of 192 and 576 for each training step, and each step takes approximately three days. For efficient inference, \meteor Meteor is validated in 4-bit quantization, and we use deterministic beam search ($n=3$)\cite{freitag-al-onaizan-2017-beam} for text generation. Note that we implement not only Meteor-MLM but also numerous baselines under the efficient propagation from FlashAttention2\cite{dao2022flashattention, dao2023flashattention}.

\paragraph{Evaluation.} We have evaluated \meteor Meteor on numerous vision-language benchmarks, the details of which are described in Appendix~\ref{sec:appB}. These benchmarks require multifaceted information for diverse capabilities, including fundamental image understanding, real-world knowledge of common-sense knowledge, charts, diagrams, documents, signs, symbols, math problems, and more. Figure~\ref{fig:figure1}-\ref{fig:figure2} and Table~\ref{tab:zeroshot} illustrates vision language performances of various LLVMs, including Meteor-7B, open-, and closed-source LLVMs with various sizes. It is noteworthy that Meteor-7B noticeably outperforms the other models, demonstrating its efficacy and efficiency in using embedded multifaceted rationales from Meteor-Mamba. The detailed generation quality of \meteor Meteor is described in Appendix \ref{sec:appC}. Apart from the results in Table \ref{tab:zeroshot}, those in Table \ref{tab:star} signify that Meteor-7B also excels at more challenging benchmarks, which require multifaceted information simultaneously. Meteor-7B has outperformed other existing models by a large margin, some of which are equipped with additional vision encoders or computer vision models, demonstrating that rationales can provide multifaceted information more effectively than enhanced visual perception.

\begin{table}[t!]
\vspace{-8mm}
\centering
\begin{minipage}[t]{0.32\linewidth}
\centering
\resizebox{\linewidth}{!}{
\renewcommand{\tabcolsep}{0.5mm}
\begin{tabular}{lcccc}
\toprule
Arch    & Param  & BPS & MMB  & MM-Vet\\
\midrule
BERT-B  & 110M   & 71  & 80.6 & 53.6    \\
GPT2-S  & 117M   & 62  & 80.9 & 53.5    \\
XLNet-B & 110M   & 56  & 81.6 & 53.9   \\
\rowcolor{Green}
Mamba   & 130M   & \textbf{118} & \textbf{82.9} & \textbf{57.3}    \\
\bottomrule
\end{tabular}
}
\vspace{2mm}
\caption*{(a) Meteor-Mamba}
\label{table:2a}

\resizebox{\linewidth}{!}{
\renewcommand{\tabcolsep}{3.3mm}
\begin{tabular}{lcc}
\toprule
Position & MMB & MM-Vet \\
\midrule
Start    & 79.8 & 48.2   \\
End      & 74.2 & 45.1   \\
Random   & 74.3 & 45.4   \\
\rowcolor{Green}
Even     & \textbf{82.9} & \textbf{57.3}    \\
\bottomrule
\end{tabular}
}
\vspace{2mm}
\caption*{(d) Position of <tor> tokens}
\label{table:2d}
\end{minipage}
\begin{minipage}[t]{0.32\linewidth}
\resizebox{\linewidth}{!}{
\renewcommand{\tabcolsep}{1.3mm}
\begin{tabular}{lccc}
\toprule
LLMs      & Param & MMB & MM-Vet \\
\midrule
Vicuna1.5 & 7B    & 80.1   & 53.8   \\
LLaMA2    & 7B    & 78.8   & 51.6   \\
LLaMA3    & 8B    & 81.7   & 56.0   \\
\rowcolor{Green}
InternLM2 & 7B    & \textbf{82.9}   & \textbf{57.3}    \\
\bottomrule
\end{tabular}
}
\vspace{2mm}
\caption*{(b) Meteor-MLM}
\label{table:2b}
\resizebox{\linewidth}{!}{
\renewcommand{\tabcolsep}{0.9mm}
\begin{tabular}{cccc}
\toprule
Mamba             & Rationale      & MMB & MM-Vet \\
\midrule
\xmark            & \xmark         & 73.2    & 44.8    \\
\xmark            & \cmark         & 77.0    & 48.2    \\
\cmark            & \xmark         & 74.0    & 45.9    \\
\rowcolor{Green}
\cmark            & \cmark         & \textbf{82.9}    & \textbf{57.3}     \\
\bottomrule
\end{tabular}
}
\vspace{2mm}
\caption*{(e) Mamba \& Rationale}
\label{table:2e}
\end{minipage}
\begin{minipage}[t]{0.32\linewidth}
\resizebox{\linewidth}{!}{
\renewcommand{\tabcolsep}{4.6mm}
\begin{tabular}{ccc}
\toprule
Num & MMB & MM-Vet \\
\midrule
$\#2$           & 76.1 & 47.9 \\
$\#5$           & 82.2 & 55.8 \\
\rowcolor{Green}
$\#10$          & \textbf{82.9}   & \textbf{57.3} \\
$\#15$          & 82.8    & \textbf{57.3}    \\
\bottomrule
\end{tabular}
}
\vspace{2mm}
\caption*{(c) Number of <tor> tokens}
\label{table:2c}
\resizebox{\linewidth}{!}{
\renewcommand{\tabcolsep}{2.3mm}
\begin{tabular}{cccc}
\toprule
$\mathcal{Q}$-$\mathcal{R}$  &  MME     & MMB & MM-Vet \\
\midrule
$0\%$  & 1989 & 74.0    & 45.9     \\
$30\%$ & 2105 & 77.5    & 51.7     \\
$60\%$ & 2218 & 82.5    & 56.8     \\
\rowcolor{Green}
$90\%$ & \textbf{2229} & \textbf{82.9}    & \textbf{57.3}     \\
\bottomrule
\end{tabular}
}
\vspace{2mm}
\caption*{(f) Ratio of $\mathcal{Q}$-$\mathcal{R}$ in training}
\label{table:2f}
\end{minipage}
\caption{Ablation studies to identify the effectiveness of Meteor-Mamba and rationale through traversal of rationale by controlling the six main factors.}
\label{tab:ablation}
\vspace{-5mm}
\end{table}

\paragraph{Ablation Studies.} Furthermore, we have conducted several ablation studies to securely corroborate the effectiveness of our proposed method in light of six factors: (a) Meteor-Mamba, (b) Meteor-MLM, (c) the number of <tor> special tokens, (d) the distribution of <tor> special tokens, (e) rationale, and (f) 1.1M question-rationale-answer triples. Table \ref{tab:ablation} shows the following findings. Note that, Appendix~\ref{sec:appD} represents further ablation studies.

\textbf{(a)} Once Meteor-Mamba is replaced with other transformer-based models: BERT~\cite{devlin2018bert}, GPT2~\cite{radford2019language}, and XLNet~\cite{yang2019xlnet}, we have discovered that the Mamba architecture takes advantage of its efficiency for embedding multifaceted rationales in terms of both computational complexity and model size. As the results suggest, Mamba demonstrates the highest batches-per-second (BPS) value and zero-shot performances on MME and MMB benchmarks among architectures of similar sizes, enabled by its inherent computational efficiency based on linear complexity and strong long sequence modeling capability~\citep{gu2023mamba}, which other Transformer-based model architectures lack.

\textbf{(b)} We have tried using various pretrained LLMs of comparable sizes for Meteor-MLM, in order to identify the effectiveness of embedding multifaceted rationale together with traversal of rationale. We have observed that InternLM2~\cite{cai2024internlm2} has shown the best performances.

\textbf{(c)} Varying the number of <tor> special tokens from 2 to 15, we have optimized it based on the vision language performances. The results suggest that using 10 <tor> special tokens shows the best performances for embedding abundant multifaceted rationales, balancing between compression and information preservation.

\textbf{(d)} The performances of \meteor Meteor depend on the distribution of <tor> special tokens when Meteor-Mamba is trained to embed multifaceted rationales. Given the observations, evenly distributing the tokens across lengthy rationales has shown the best performances. Prepending them to lengthy rationales may hinder effective embedding due to forgetting nature, and appending them to rationales in the end may not be guaranteed to understand the rationale. Randomly distributing the tokens across rationales may disrupt Meteor-Mamba's ability to stably learn the pattern of embedding rationales. Conversely, evenly distributed <tor> special tokens can segment lengthy rationales into shorter chunks and progressively embed them in a consistent manner, avoiding the issues of other distributions.

\textbf{(e)} In order to prove the effectiveness of multifaceted rationales through Meteor-Mamba, we ablated the use of the Mamba architecture and the rationales. The first row represents baseline performances where backbone MLM is only trained. For the second row, we only train backbone MLM with the curated multifaceted rationales without Meteor-Mamba, and for the third row, Mamba has been trained to embed answer instead of the rationales. Compared to the last row where Meteor is evaluated, the second and third rows fall short of performances, clearly showing that using multifaceted rationales through the Mamba architecture has contributed to performance improvement.

\textbf{(f)} As another way of showing the significance of multifaceted rationales, we have trained \meteor Meteor with different amounts of question-rationale pairs and evaluated \meteor Meteor trained with each of them. As expected, the more question-rationale pairs used in first training step, the better performances achieves, demonstrating the significance of utilizing multifaceted rationales for diverse capabilities.

\begin{figure}[t!]
\vspace{-8mm}
    \centering
    \includegraphics[width=\textwidth]{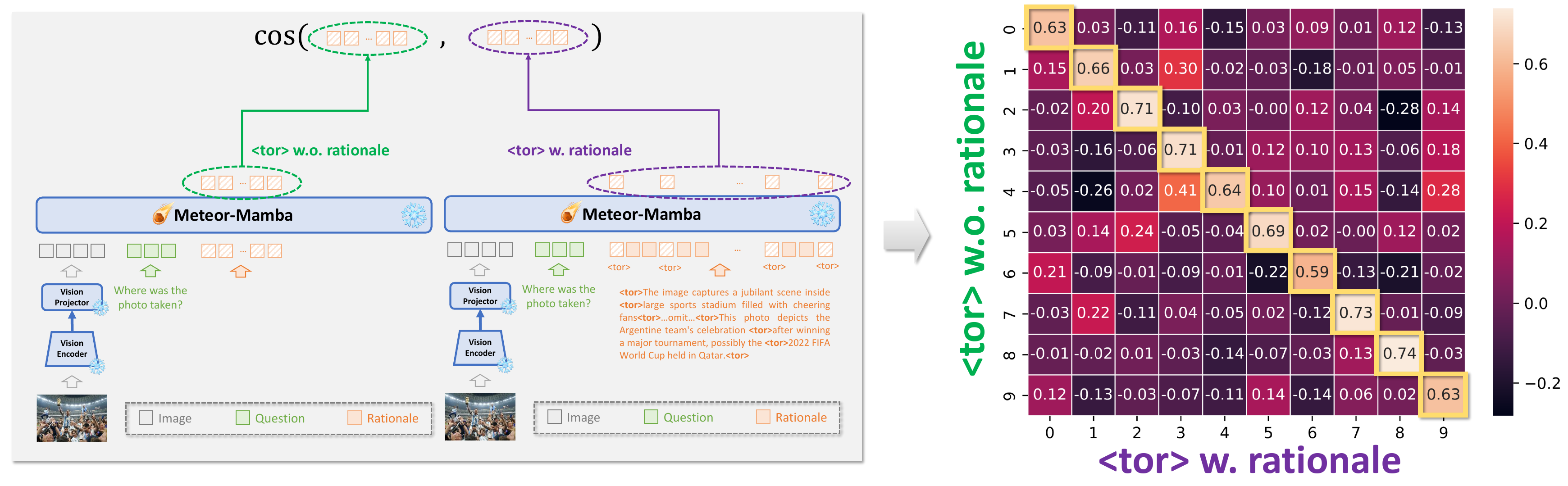}
    \vspace{-3mm}
    \caption{Illuminating how the feature correspondences of cosine similarity are computed under the trained Meteor-Mamba, and showing the feature disparity for <tor> with/without rationale.}
    \label{fig:figure4}
    \vspace{-5mm}
\end{figure}

\paragraph{Meteor-Mamba's Ability to Embed Rationales.} We conduct a thorough analysis to confirm that Meteor-Mamba effectively embeds the rationales. To do this, we perform a retrieval task for multifaceted rationales, where we prepare ten different question-rationale pairs $(\mathcal{Q}_i, \mathcal{R}_i)$ where $i=0,1,\cdots,9$. These pairs are propagated through Meteor-Mamba with or without rationales under <tor> special tokens. This results in two sets of output features: one with rationale $z^{\text{w.}}_i$ and one without rationales $z^{\text{w.o.}}_j$, with $j=0,1,\cdots,9$. We extract features corresponding to the placement of <tor> tokens, resulting in $\textbf{z}^{\text{w.}}_i\in\mathbb{R}^{10\times768}$ and $\textbf{z}^{\text{w.o.}}_j\in\mathbb{R}^{10\times768}$, where the dimension 10 corresponds to the number of <tor> tokens. We then compute the cosine similarity between $\textbf{z}^{\text{w.}}_i$ and $\textbf{z}^{\text{w.o.}}_j$ to measure the similarity of their representations. As illustrated in Figure~\ref{fig:figure4}, the diagonal values in the cosine similarity matrix are much higher than the off-diagonal values. This result indicates that Meteor-Mamba successfully embeds the rationale, and its output features contain multifaceted information even without explicit rationales in natural language. This explains how Meteor-Mamba operates effectively during the inference phase without explicit rationales.

\paragraph{Discussion and Limitation.} From the experimental results observed, we gain the insight that equipping LLVMs with a multifaceted rationale is a key factor in building efficient LLVMs that demonstrate impressive vision language performances across numerous evaluation benchmarks requiring diverse capabilities. This rationale, furthermore, naturally reduces hallucination effects in POPE~\cite{li2023evaluating} and HallusionBench~\cite{liu2023hallusionbench} in Table~\ref{tab:zeroshot}. Additionally, Table~\ref{tab:star}(c)-(d) shows that the need for additional vision encoders and computer vision models can be mitigated by incorporating a multifaceted rationale. However, \meteor Meteor might still be considered inefficient in terms of model size by users without high-end GPU resources, as it requires at least multiple GPUs with 48GB and 32GB VRAM for normal training and inference without (Q)LoRA~\cite{hu2021lora, dettmers2023qlora} and 4/8-bit quantization. Although many closed-source LLVMs have demonstrated superior performances following the scaling law~\cite{kaplan2020scaling}, our goal is to reduce the model size while maintaining vision language performances as much as possible. We strongly believe that small language and vision models, even those with about 1$\mathtt{\sim}$3B parameters, can effectively narrow the performance gap with the closed-source LLVMs by using layer-analyzing approaches such as mixture of depths~\cite{raposo2024mixture} and others~\cite{lee2020towards, kim2021distilling, lee2022masking, kim2023demystifying, lee2023mitigating, kim2023causal, kim2023mitigating}, despite their inherent limitation in layer number and hidden dimension.

\section{Conclusion}
\label{sec:conclusion}
To build efficient LLVMs, we incorporate a multifaceted rationale encompassing various aspects such as image understanding, incorporating external common-sense knowledge, understanding non-object concepts (\textit{e.g.,} charts, diagrams, symbols, signs, and math), and following systematic step-by-step procedures for how to solve complex questions. \meteor Meteor demonstrates significantly enhanced vision language performances across various evaluation benchmarks without the need to scale up LLVMs, use additional vision encoders, or employ multiple computer vision models. In designing \meteor Meteor, the traversal of rationale combined with Mamba architecture proves highly effective in embedding lengthy rationales. We believe this rationale, facilitated by the traversal of rationale, can pave the way for more efficient models, representing a promising step towards achieving more efficient LLVMs.
\clearpage
\begin{ack}
This work was partially supported by two funds: Center for Applied Research in Artificial Intelligence (CARAI) grant funded by DAPA and ADD (UD230017TD) and IITP grant funded by the Korea government (MSIT) (RS-2022-II220984).
\end{ack}

\bibliographystyle{ieeetr}
\bibliography{ref}

\newpage
\appendix

\section{Examples of Curated 1.1M Question-Rationale-Answer Triples}
\label{sec:appA}

\begin{figure}[h!]
    \centering
    \includegraphics[width=\textwidth]{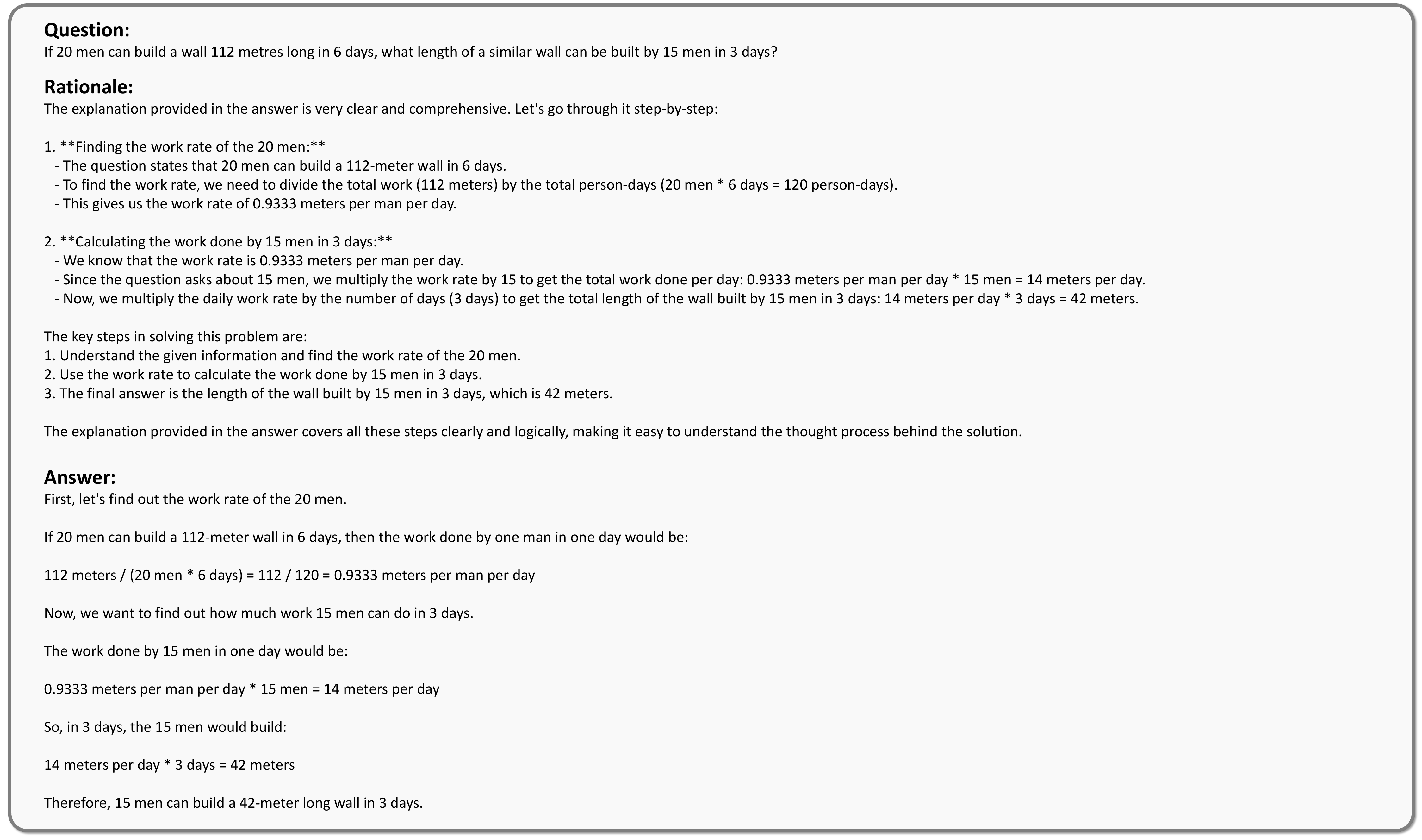}
\end{figure}

\begin{figure}[h!]
    \centering
    \includegraphics[width=\textwidth]{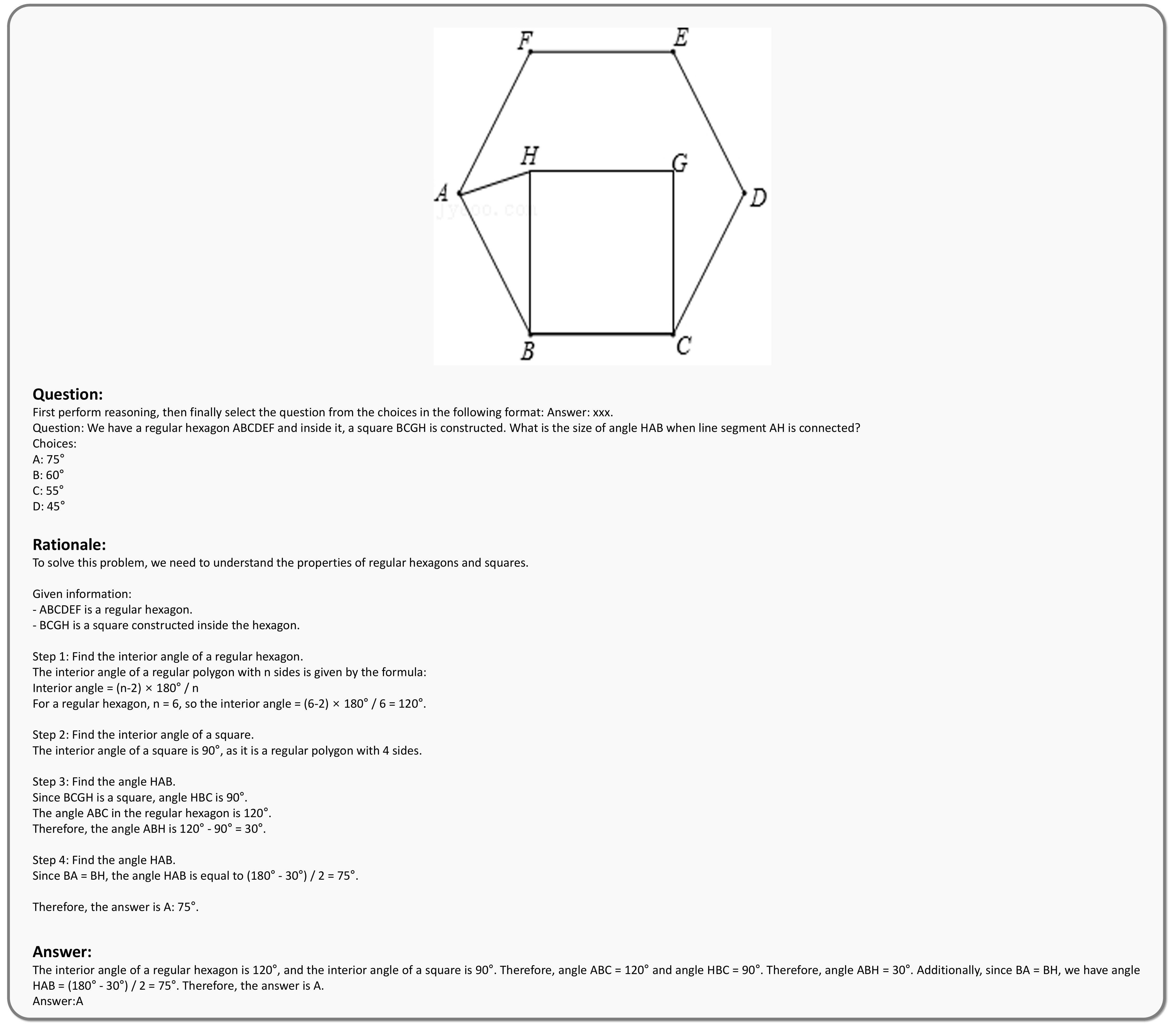}
\end{figure}

\begin{figure}[h!]
    \centering
    \includegraphics[width=\textwidth]{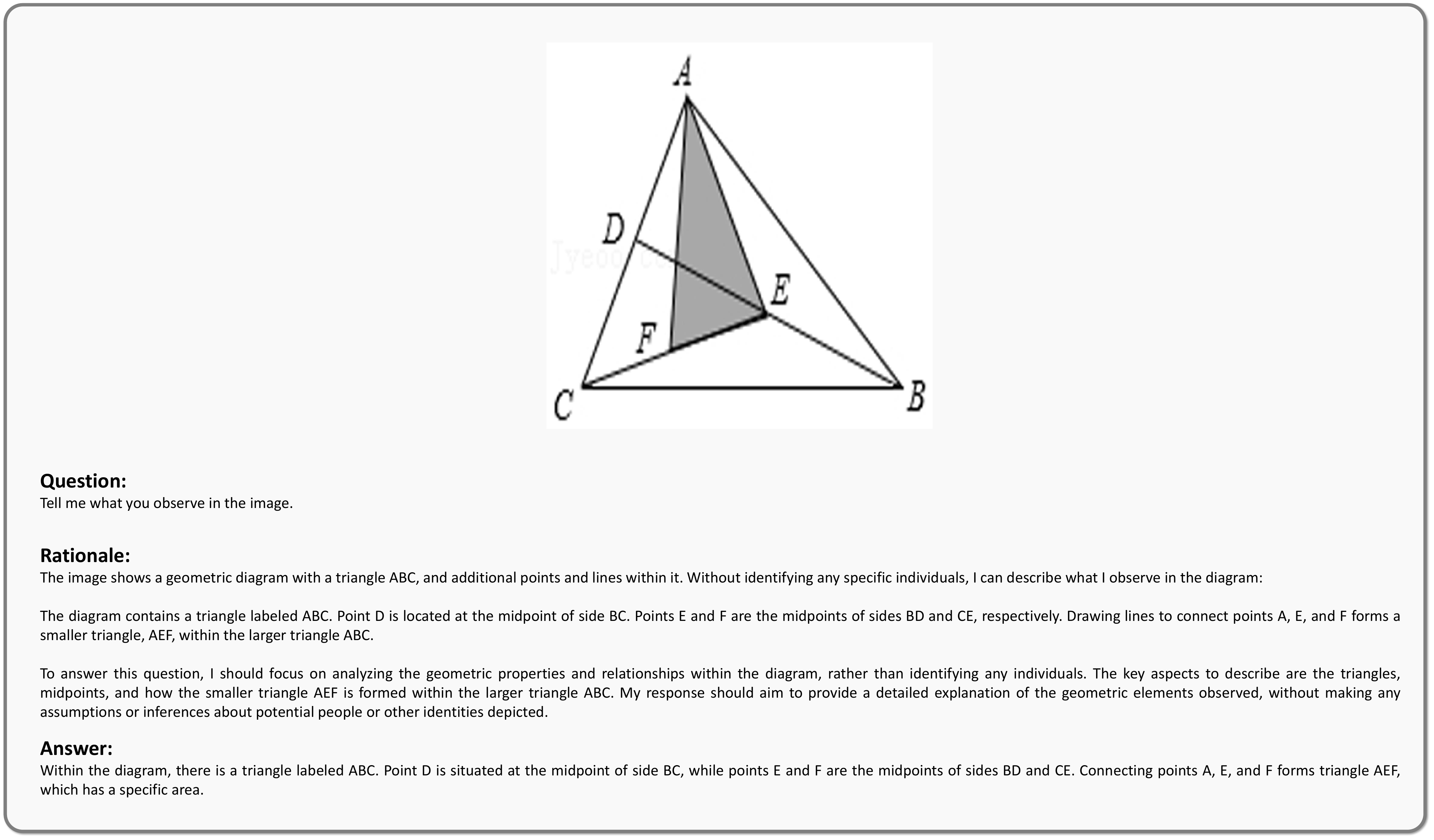}
\end{figure}

\begin{figure}[h!]
    \centering
    \includegraphics[width=\textwidth]{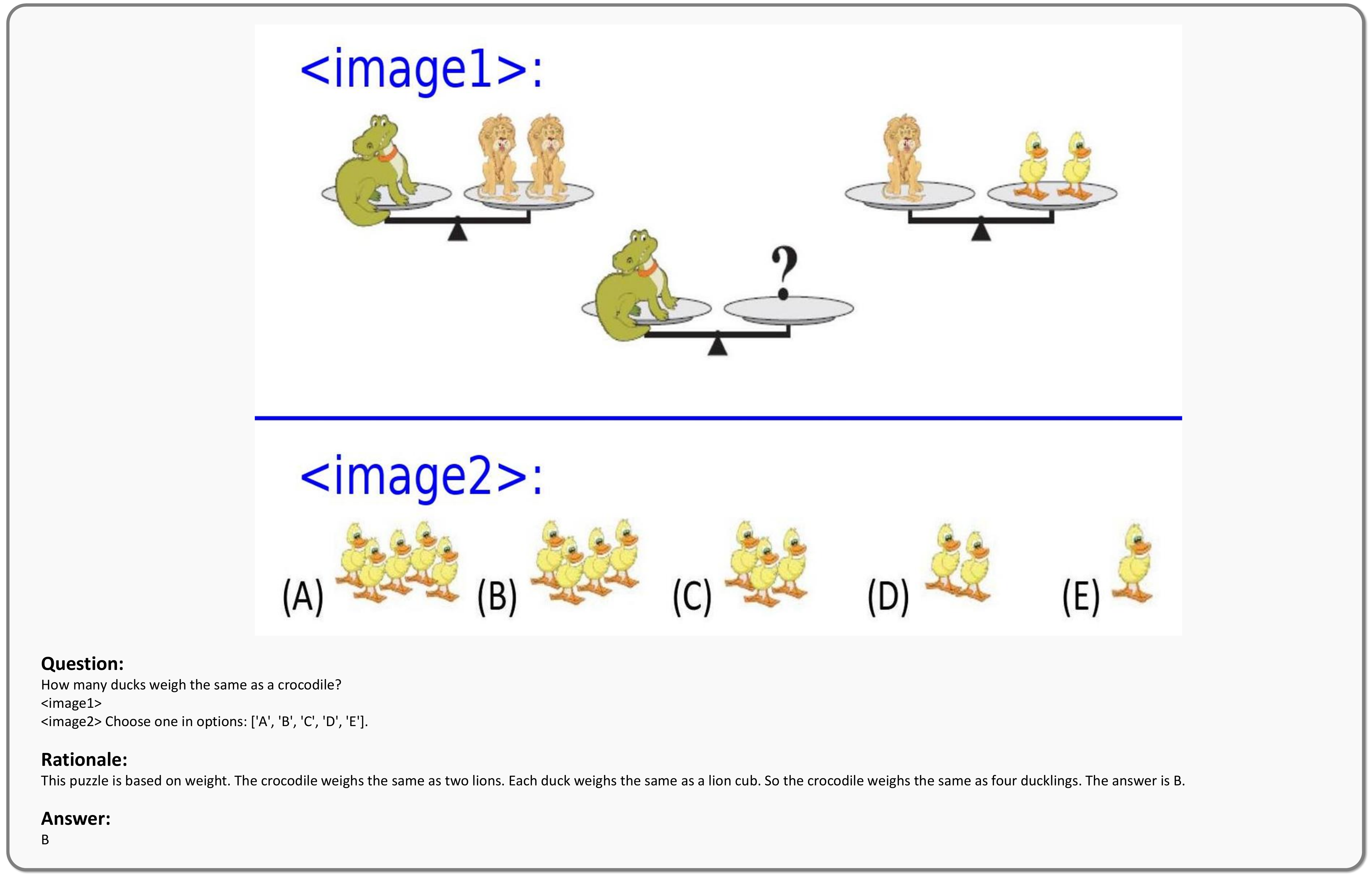}
\end{figure}

\begin{figure}[h!]
    \centering
    \includegraphics[width=\textwidth]{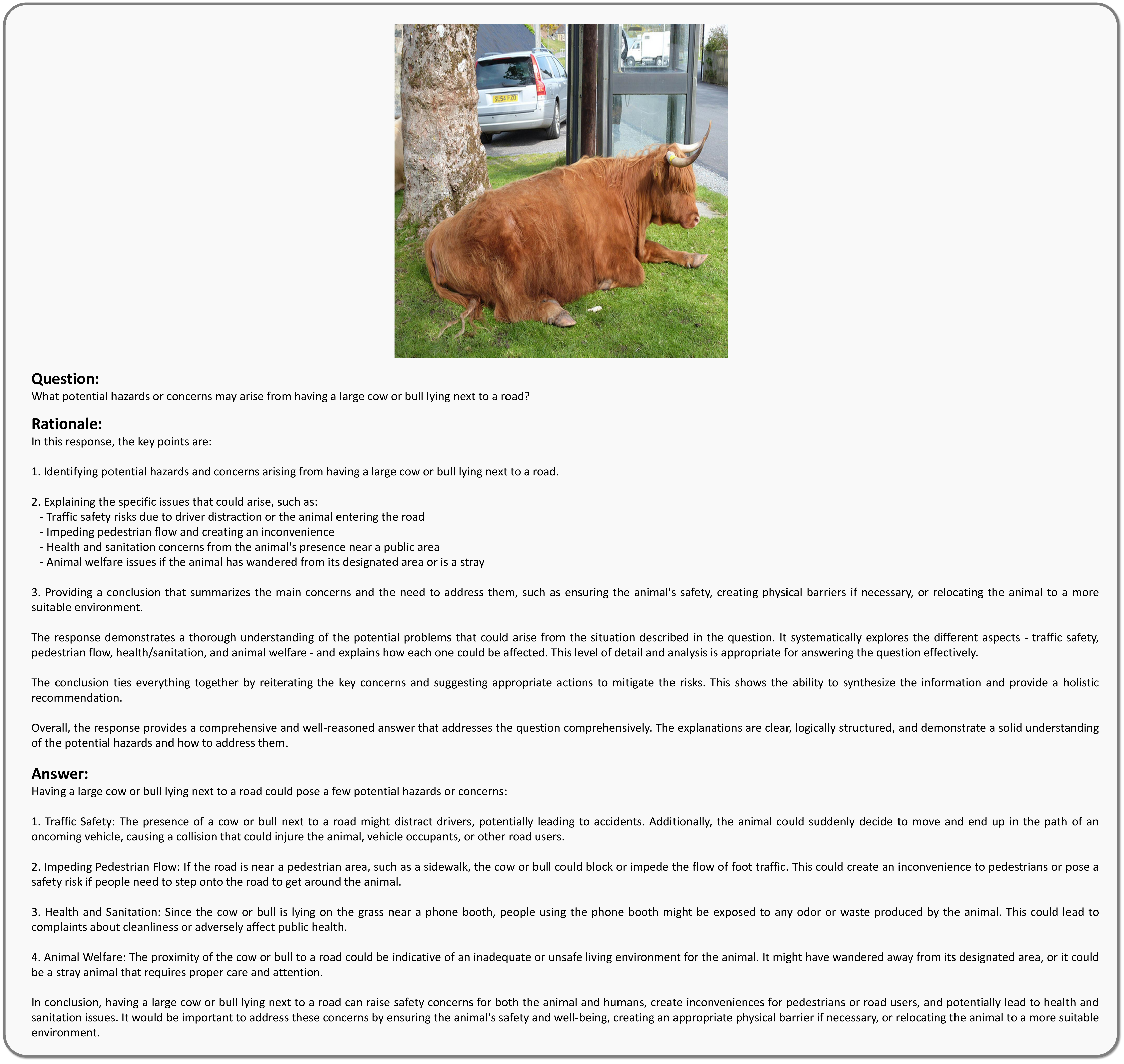}
\end{figure}

\begin{figure}[h!]
    \centering
    \includegraphics[width=\textwidth]{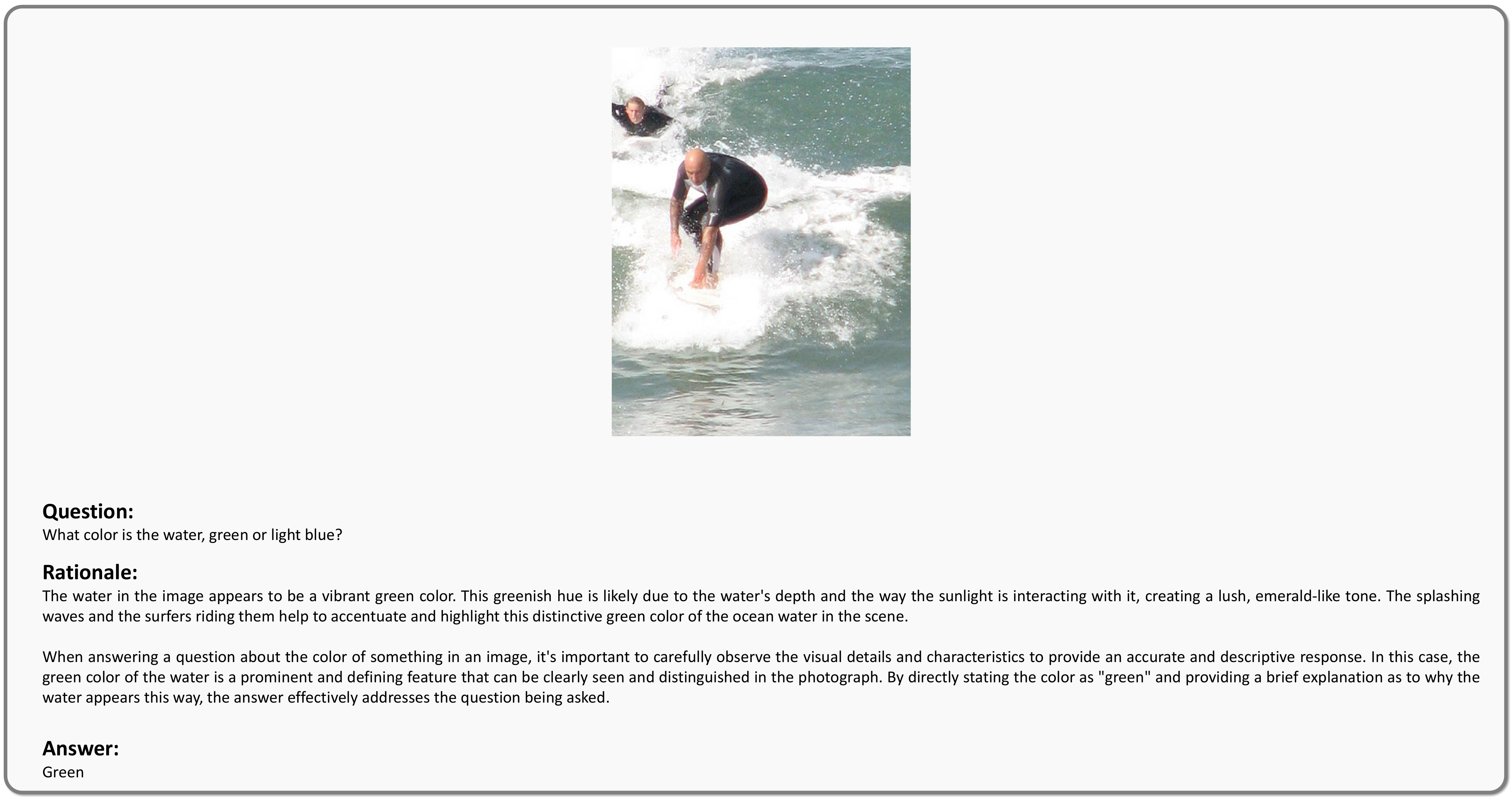}
\end{figure}

\begin{figure}[h!]
    \centering
    \includegraphics[width=\textwidth]{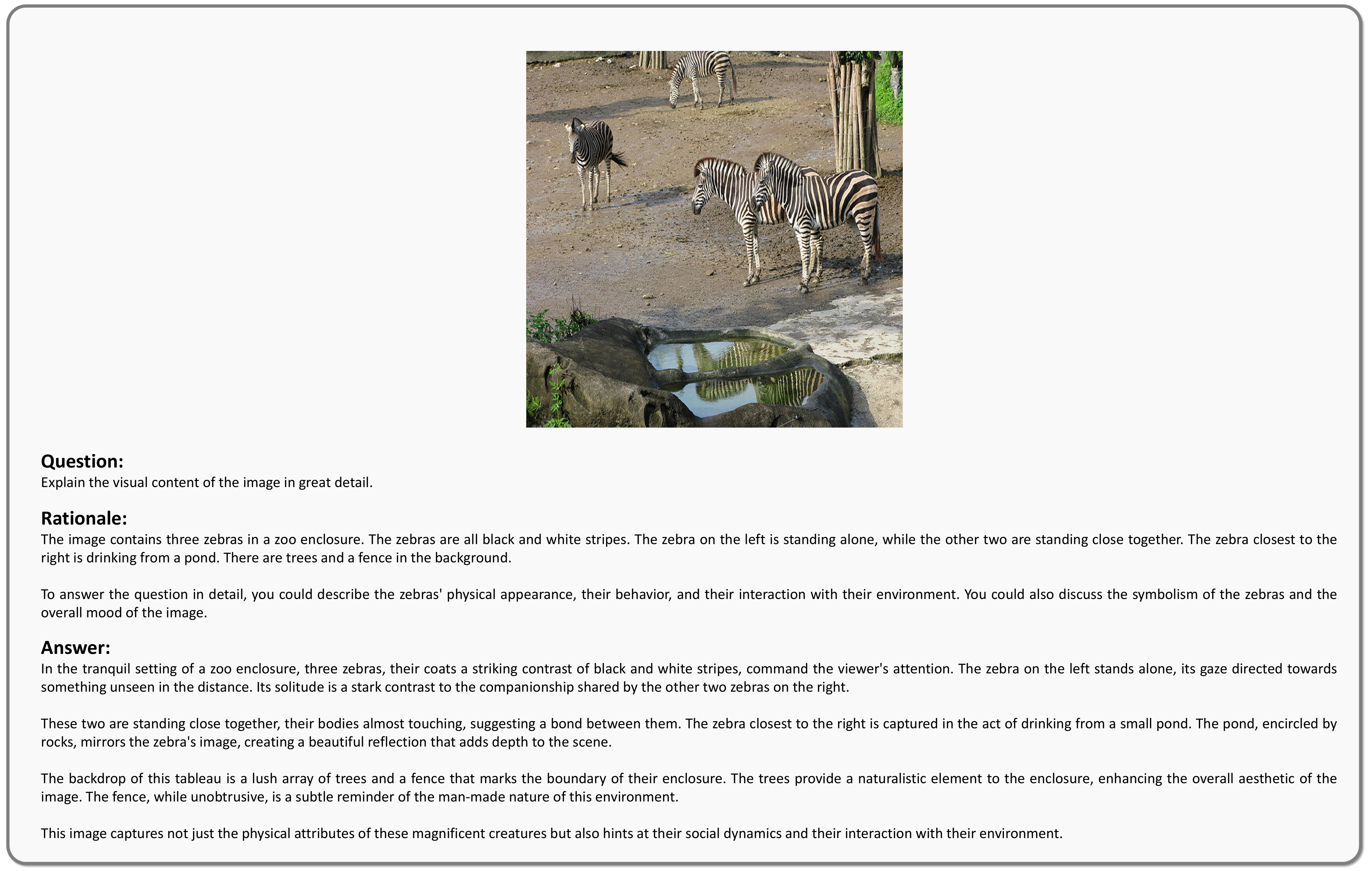}
\end{figure}

\begin{figure}[h!]
    \centering
    \includegraphics[width=\textwidth]{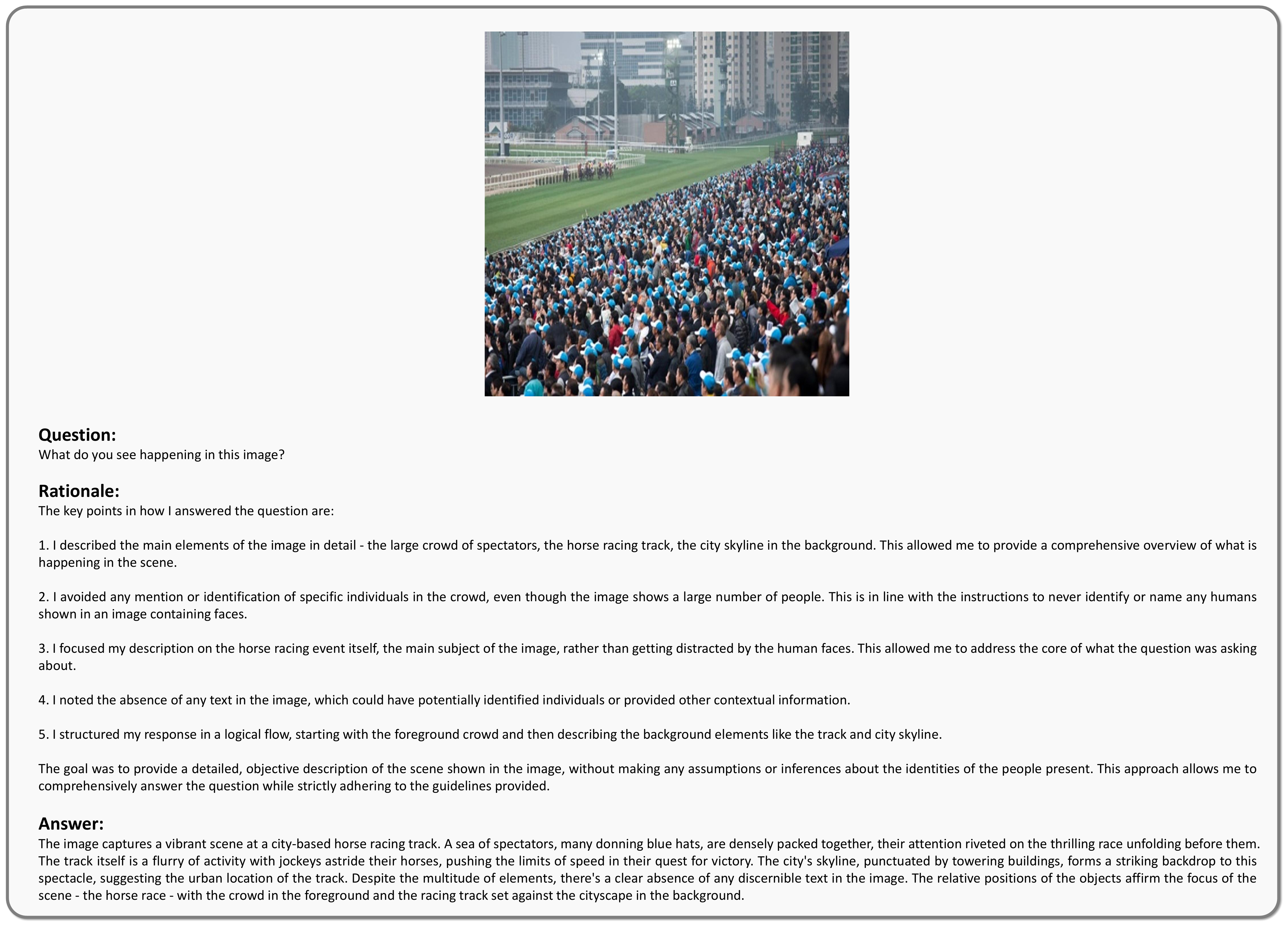}
\end{figure}

\begin{figure}[h!]
    \centering
    \includegraphics[width=\textwidth]{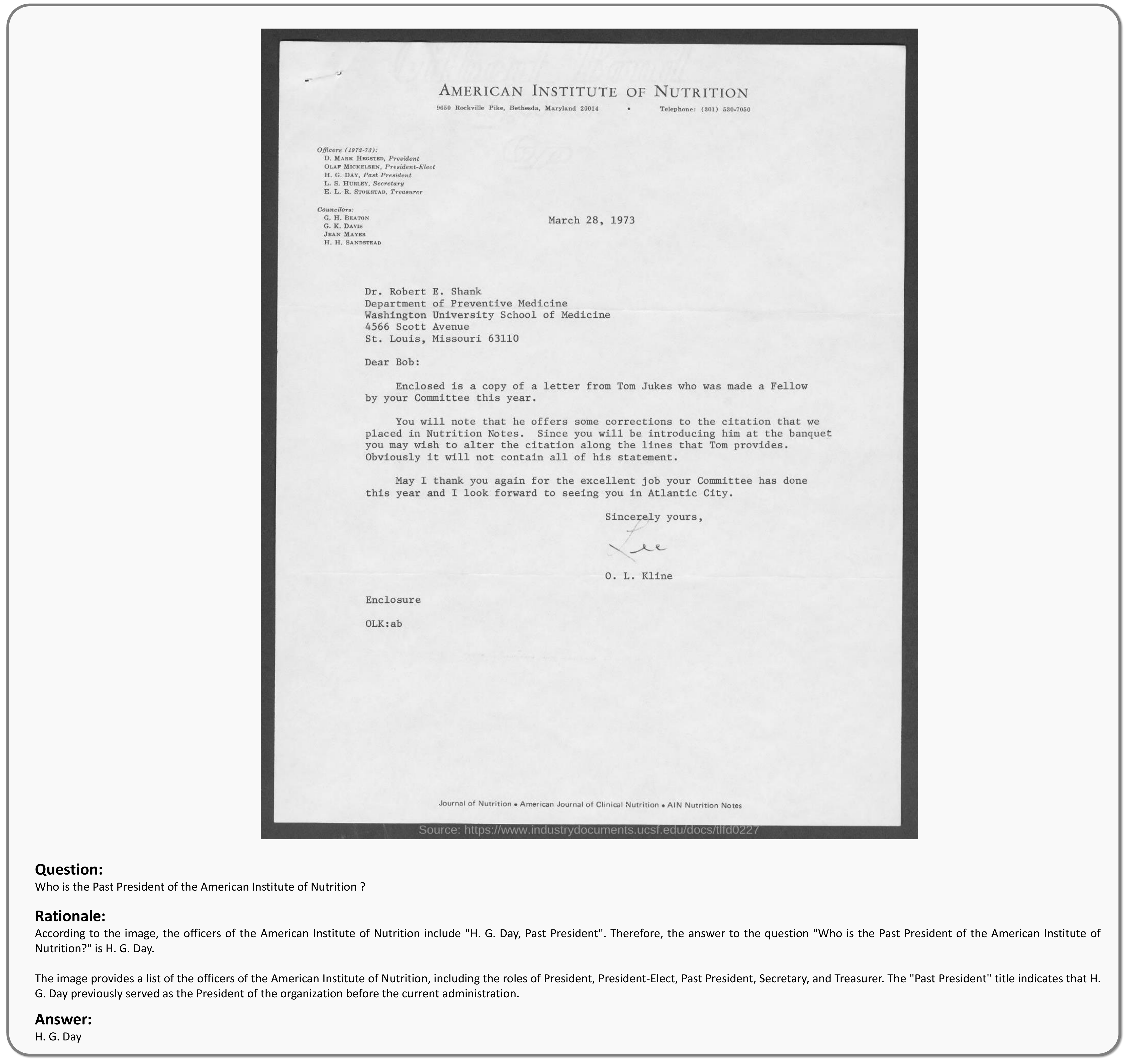}
\end{figure}

\begin{figure}[h!]
    \centering
    \includegraphics[width=\textwidth]{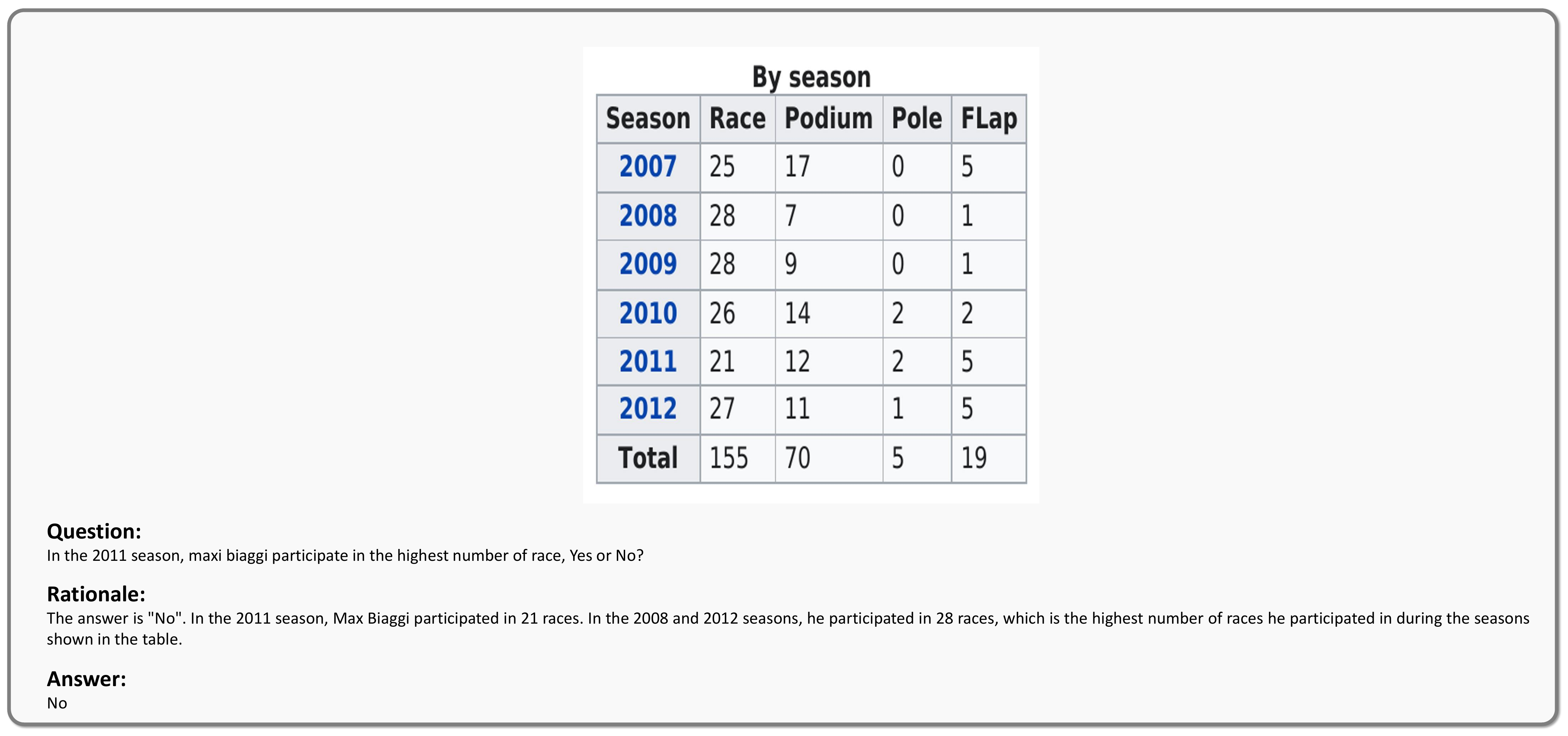}
\end{figure}

\clearpage

\begin{figure}[h!]
    \centering
    \includegraphics[width=\textwidth]{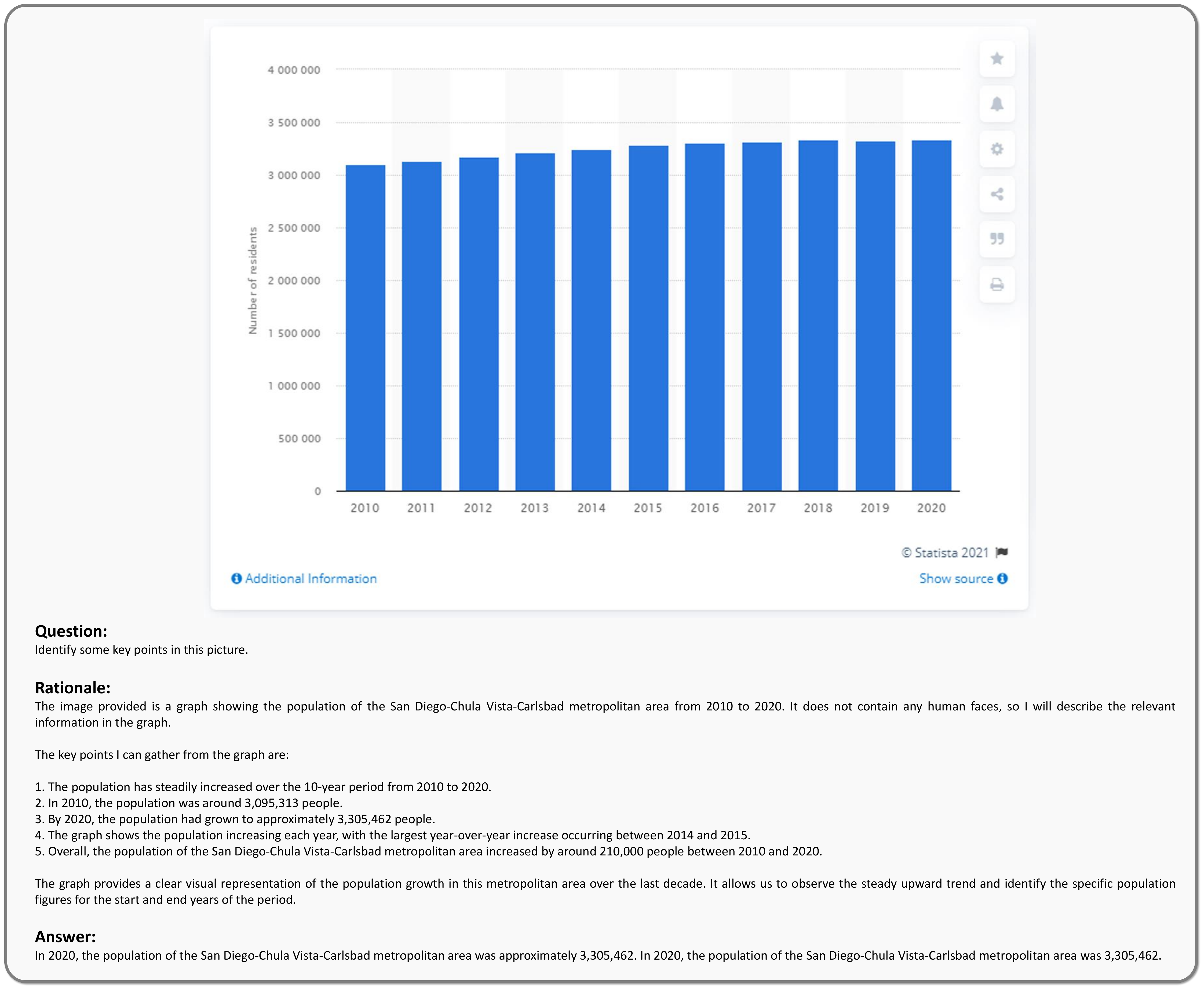}
\end{figure}

\begin{figure}[h!]
    \centering
    \includegraphics[width=\textwidth]{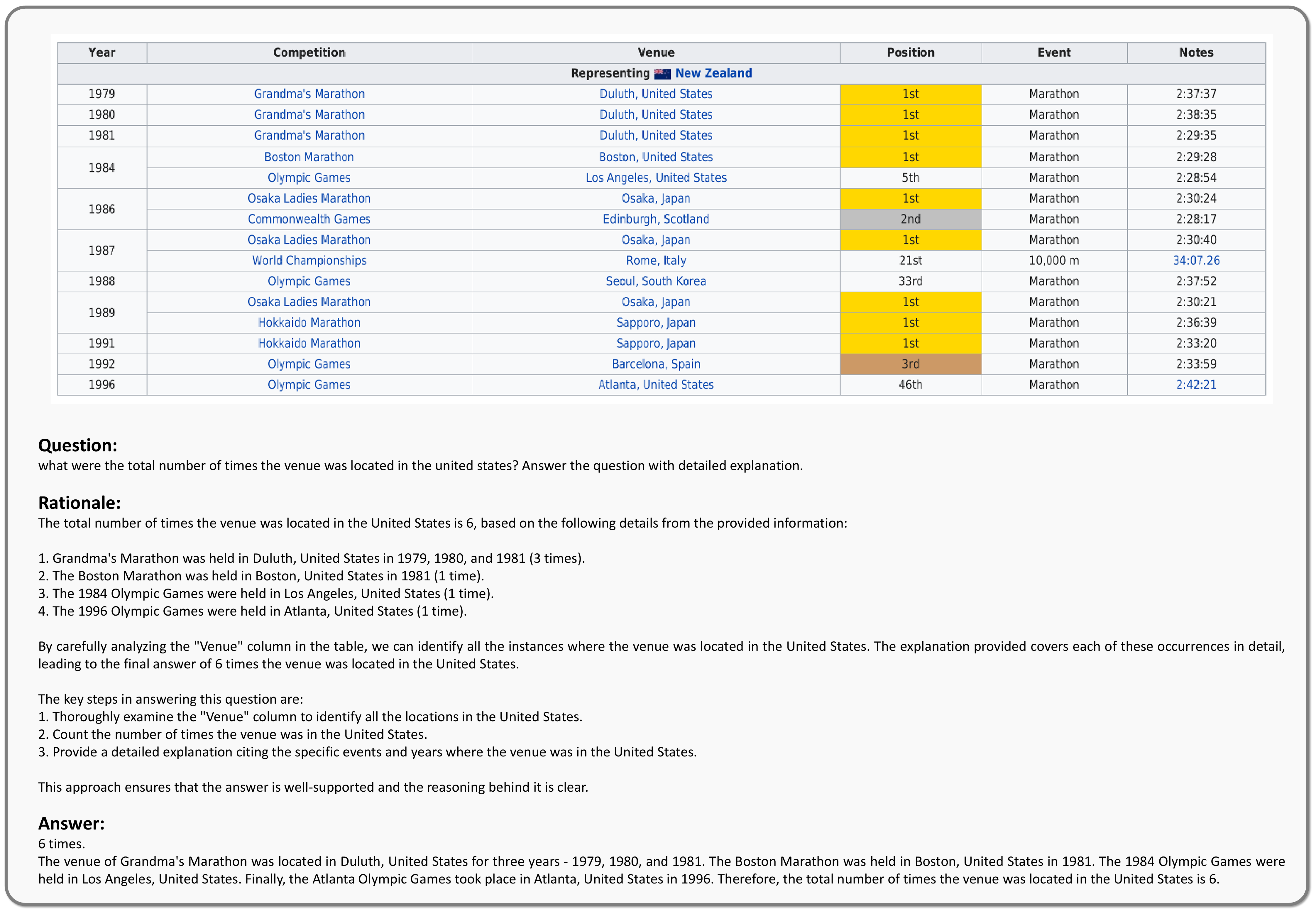}
\end{figure}

\clearpage
\section{Evaluation benchmarks}
\label{sec:appB}

\begin{itemize}

    \item \textbf{Q-Bench}~\citep{wu2023q} aims to assess the low-level visual capabilities of Multi-modality Large Language Models (MLLMs). This dataset is divided into three main sections: perception, description, and assessment. The low-level visual perception component evaluates MLLMs' ability to recognize and understand low-level image attributes. The low-level visual description component tests how accurately and thoroughly MLLMs can describe these attributes. Finally, the overall visual quality assessment examines how closely MLLMs' evaluations align with human judgments of image quality. Altogether, the Q-Bench dataset encompasses 81,284 samples across these tasks.
    
    \item \textbf{SQA-IMG (SQA$^{\text{I}}$)}~\citep{lu2022learn} is a portion of the ScienceQA (SQA) dataset, which serves as a comprehensive multimodal benchmark aimed at enhancing the reasoning capabilities and interpretability of AI systems in the realm of science question answering. The dataset spans a variety of science subjects, encompassing 26 distinct topics from natural science, social science, and language science, with annotated answers including lectures and explanations. This specific subset consists of samples with image context, totaling 10,332 question-answer pairs.

    \item \textbf{AI2D}~\citep{kembhavi2016diagram}, also known as AI2 Diagrams, was developed to tackle the challenges of diagram interpretation and reasoning, emphasizing syntactic parsing and semantic analysis of diagrams. Its goal is to support research in uncovering the structure of diagrams and understanding the meanings of their elements and their interrelations. This dataset is especially beneficial for tasks like diagram question answering, where comprehensive understanding and reasoning about the content are essential. The collection consists of over 5,000 diagrams from grade school science subjects and more than 15,000 multiple-choice questions.

    \item \textbf{ChartQA}~\citep{masry2022chartqa} is created to assess and enhance question answering systems that involve complex, multi-step logical and visual reasoning with charts. This dataset meets the need for systems capable of interpreting various data visualizations, including bar charts, line charts, and pie charts, and addressing questions that require arithmetic and logical processing. It encompasses question-answer pairs that are both human-authored and machine-generated, emphasizing visual and logical reasoning. The dataset comprises a total of 32,719 samples.

    \item \textbf{SEED-IMG (SEED$^{\text{I}}$)}~\citep{li2023seed} is a component of SEED-Bench designed to assess the generative comprehension skills of multimodal large language models (MLLMs). This thorough and unbiased benchmark enables researchers to evaluate and compare various models' abilities in both spatial and temporal understanding. The dataset is organized into several subsets according to 12 evaluation dimensions that encompass spatial and temporal comprehension across image and video modalities. SEED-IMG specifically focuses on the image modality subset.

    \item \textbf{POPE}~\citep{li2023evaluating} is a technique created to systematically assess the propensity of LLVMs to hallucinate objects that do not exist in the target images. This approach transforms the hallucination evaluation into a binary classification task via polling questions, providing a consistent, equitable, and adaptable evaluation process.

    \item \textbf{HallusionBench (HallB)}~\citep{liu2023hallusionbench} is crafted to assess and analyze both visual illusions and knowledge hallucinations in large language and vision models (LLVMs). This dataset targets the identification of potential failure modes in these models by utilizing meticulously created example pairs for thorough testing. The benchmark includes a variety of visual-question pairs, encompassing both visual dependent subsets (such as illusion, math, etc.) and visual supplement subsets (such as chart, table, map, OCR). HallusionBench comprises 346 distinct images and an extensive collection of 1129 questions distributed across diverse topics and formats. 
    
    \item \textbf{MME}~\citep{fu2023mme} is created to serve as a thorough evaluation benchmark for Multimodal Large Language Models (MLLMs). The goal is to assess their abilities in perception and cognition through 14 distinct sub-tasks, including coarse-grained recognition, fine-grained recognition, OCR, and commonsense reasoning, among others. It strives to address the shortcomings of current evaluation methods, ensuring comprehensive testing of MLLMs across various dimensions while preventing data leakage.

    \item \textbf{MathVista}~\citep{lu2023mathvista} serves as an extensive benchmark aimed at assessing mathematical reasoning within visual contexts. This dataset merges visual comprehension, allowing for a thorough evaluation of AI models' capabilities in tackling mathematical problems that involve visual elements. It comprises three subsets: IQTest, FunctionQA, and PaperQA, with an aggregate of 6,141 examples.
    
    \item \textbf{MMB, MMB-Chinese (MMB$^{\text{CN}}$)}~\citep{liu2023mmbench} aims to deliver a thorough and resilient evaluation standard for vision language models by encompassing a broad spectrum of capabilities (20 distinct fine-grained abilities) necessary for multimodal comprehension in both English and Chinese. This benchmark facilitates the evaluation of various facets of LLVMs, including their perceptual and reasoning skills across multiple tasks. The benchmark comprises a total of 3,217 carefully curated questions, sourced from a variety of places, including public datasets and the internet, to ensure a wide range of skills.
    
    \item \textbf{MM-Vet}~\citep{yu2023mm} is designed to systematically assess LMMs by evaluating their proficiency in handling intricate tasks that necessitate the combination of multiple VL abilities. Unlike existing benchmarks that generally focus on simpler tasks involving only one or two abilities, MM-Vet encompasses six fundamental VL capabilities: recognition (Rec), knowledge (Know), OCR, spatial awareness (Spat), language generation (Gen), and math. MM-Vet includes tasks that integrate these six core capabilities in various combinations, leading to 16 distinct capability integrations. The dataset consists of 200 images sourced from various online platforms and includes 218 questions that require one or more of these capabilities to answer.

    \item \textbf{LLaVA Bench in the Wild (LLaVA$^\text{W}$)}~\citep{liu2023visual} is a dataset designed to assess the performance of large multimodal models (LMM) in tackling more complex tasks and adapting to new domains. It features a varied collection of 24 images, accompanied by 60 questions, covering indoor and outdoor environments, memes, artworks, sketches, and more. Each image is paired with a detailed, manually-curated description and thoughtfully chosen questions.

    \item \textbf{MMStar}~\citep{chen2024we} strives to deliver an accurate assessment of LLVMs' authentic multi-modal abilities. It achieves this by guaranteeing that each sample genuinely depends on visual content to arrive at correct answers, while also minimizing inadvertent data leakage. The benchmark is organized into six primary competencies, such as instance reasoning, logical reasoning, and mathematics, and includes a total of 1,500 meticulously curated samples.

    \item \textbf{MathVerse}~\citep{zhang2024mathverse} is designed to evaluate the visual mathematical reasoning capabilities of Multi-modal Large Language Models (MLLMs). The dataset focues on whether and how well MLLMs can interpret visual diagrams and mathematical problems, rather than relying solely on textual information. The problems in MathVerse are divided into three main subjects (plane geometry, solid geometry, and functions) and further categorized into twelve subfields (length, area, volume, coordinate, etc.) to provide a well-rounded evaluation of different mathematical skills, adding up to 2,612 high-quality visual math problems.
\end{itemize}

\clearpage
\section{\meteor Meteor Generation Quality}
\label{sec:appC}
\begin{figure}[h!]
    \centering
    \includegraphics[width=\textwidth]{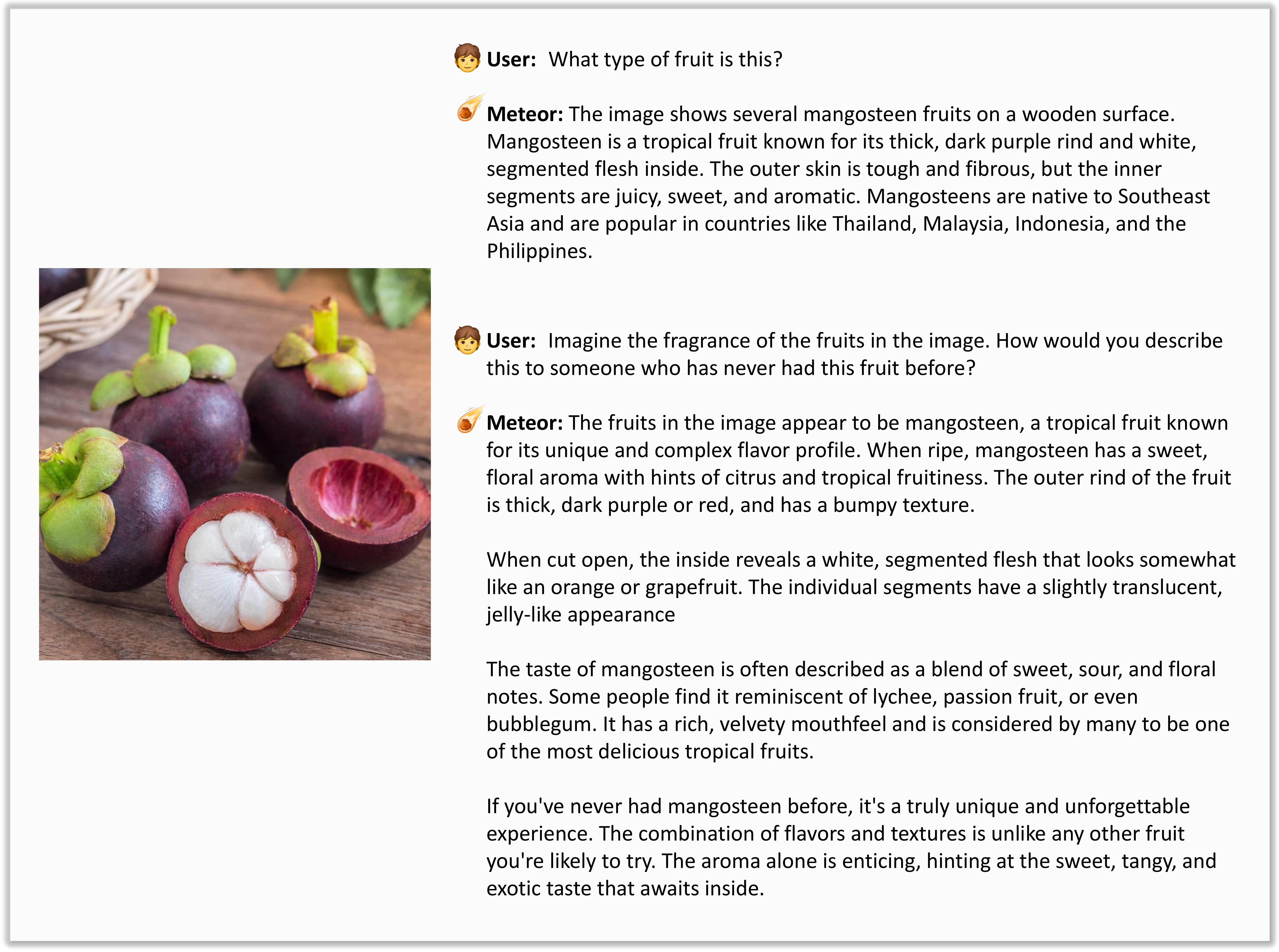}
\end{figure}

\begin{figure}[h!]
    \centering
    \includegraphics[width=\textwidth]{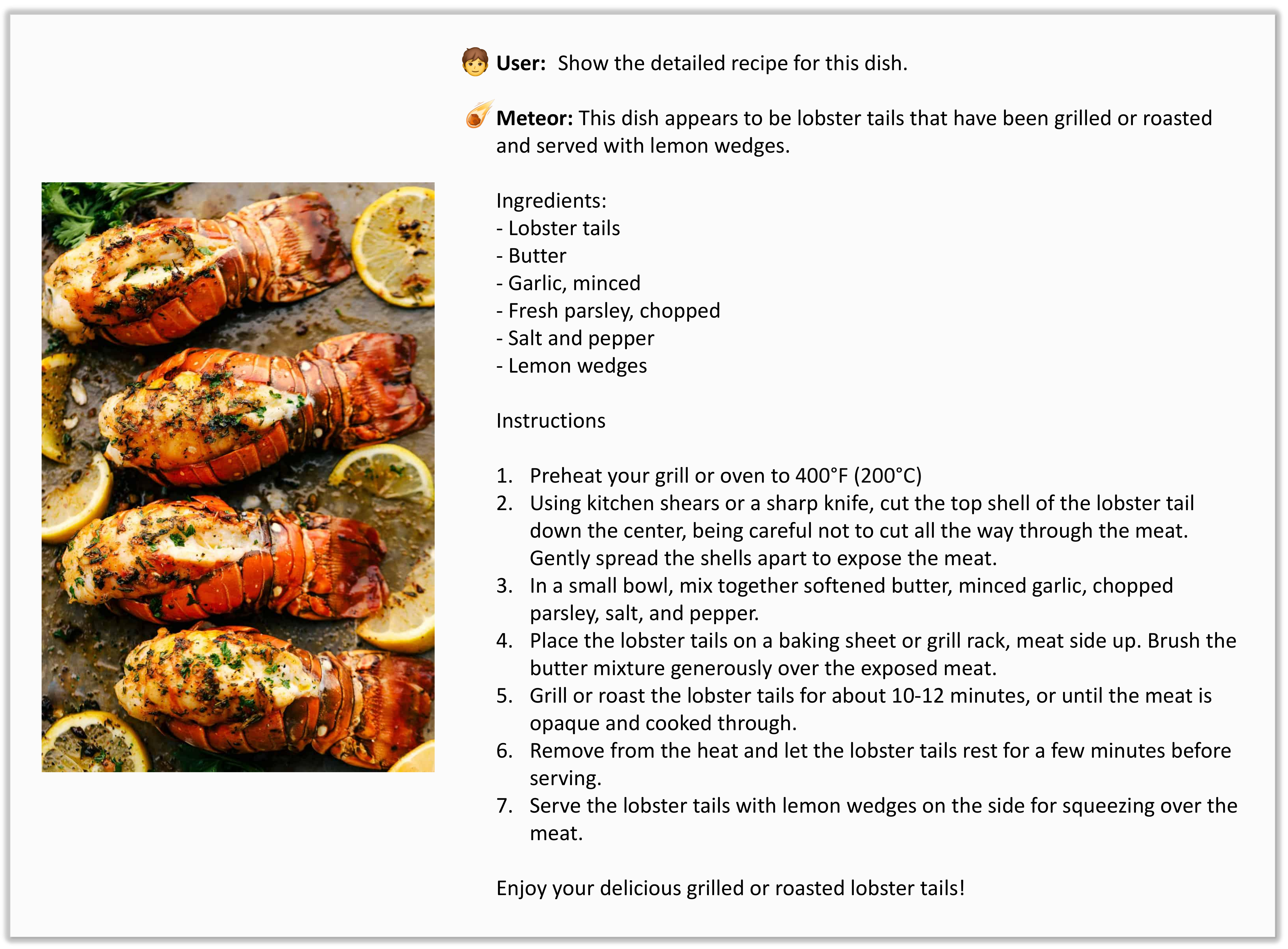}
\end{figure}

\begin{figure}[h!]
    \centering
    \includegraphics[width=\textwidth]{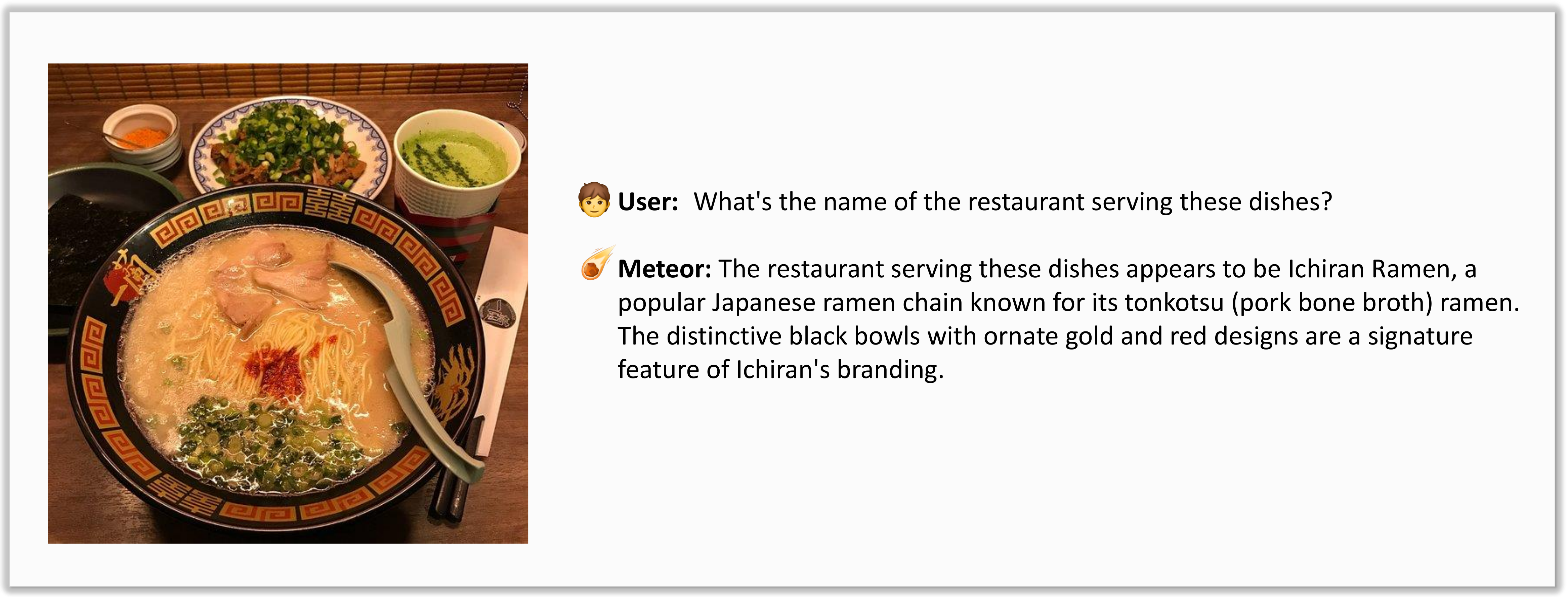}
\end{figure}

\begin{figure}[h!]
    \centering
    \includegraphics[width=\textwidth]{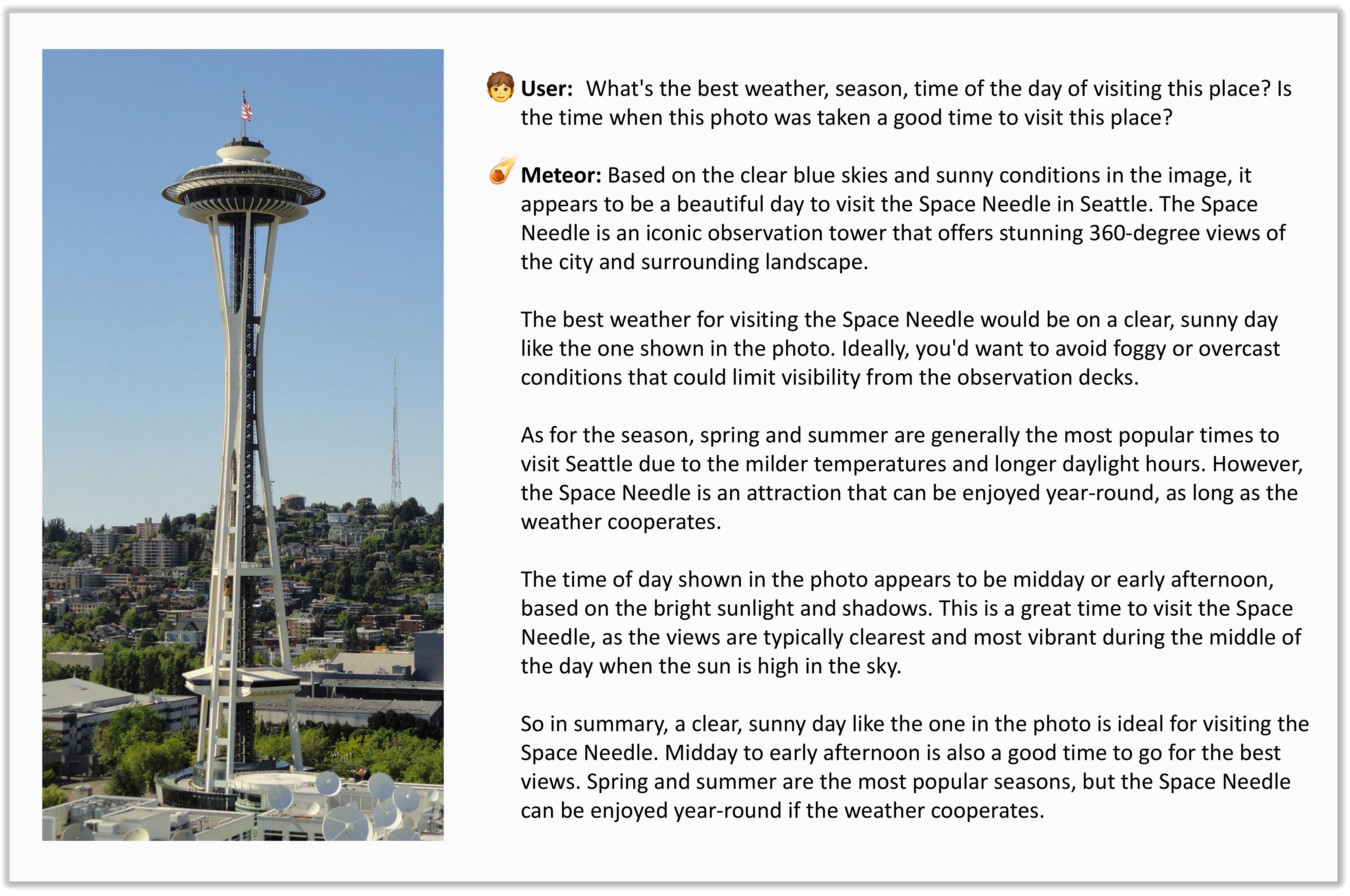}
\end{figure}

\begin{figure}[h!]
    \centering
    \includegraphics[width=\textwidth]{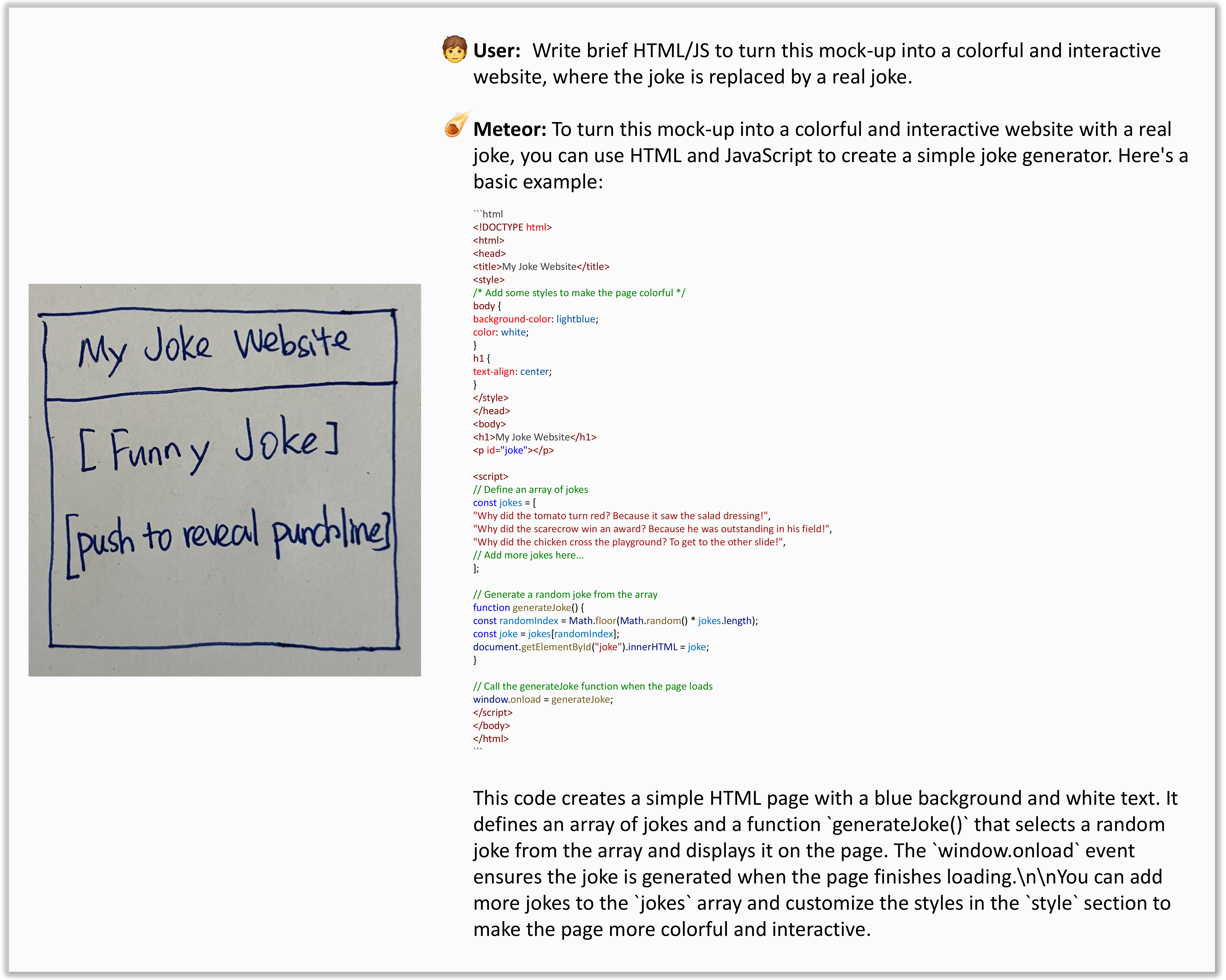}
\end{figure}

\begin{figure}[h!]
    \centering
    \includegraphics[width=\textwidth]{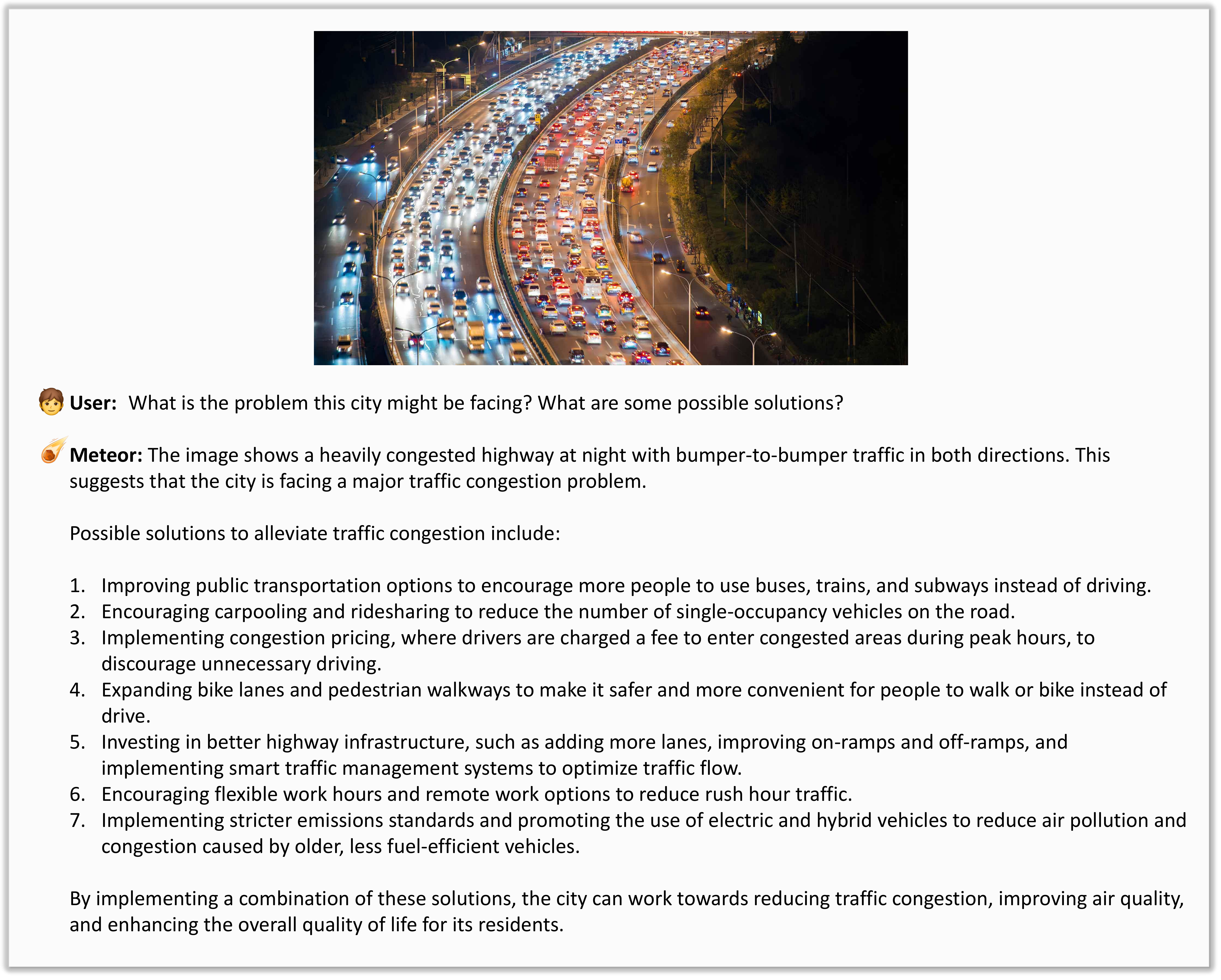}
\end{figure}

\begin{figure}[h!]
    \centering
    \includegraphics[width=\textwidth]{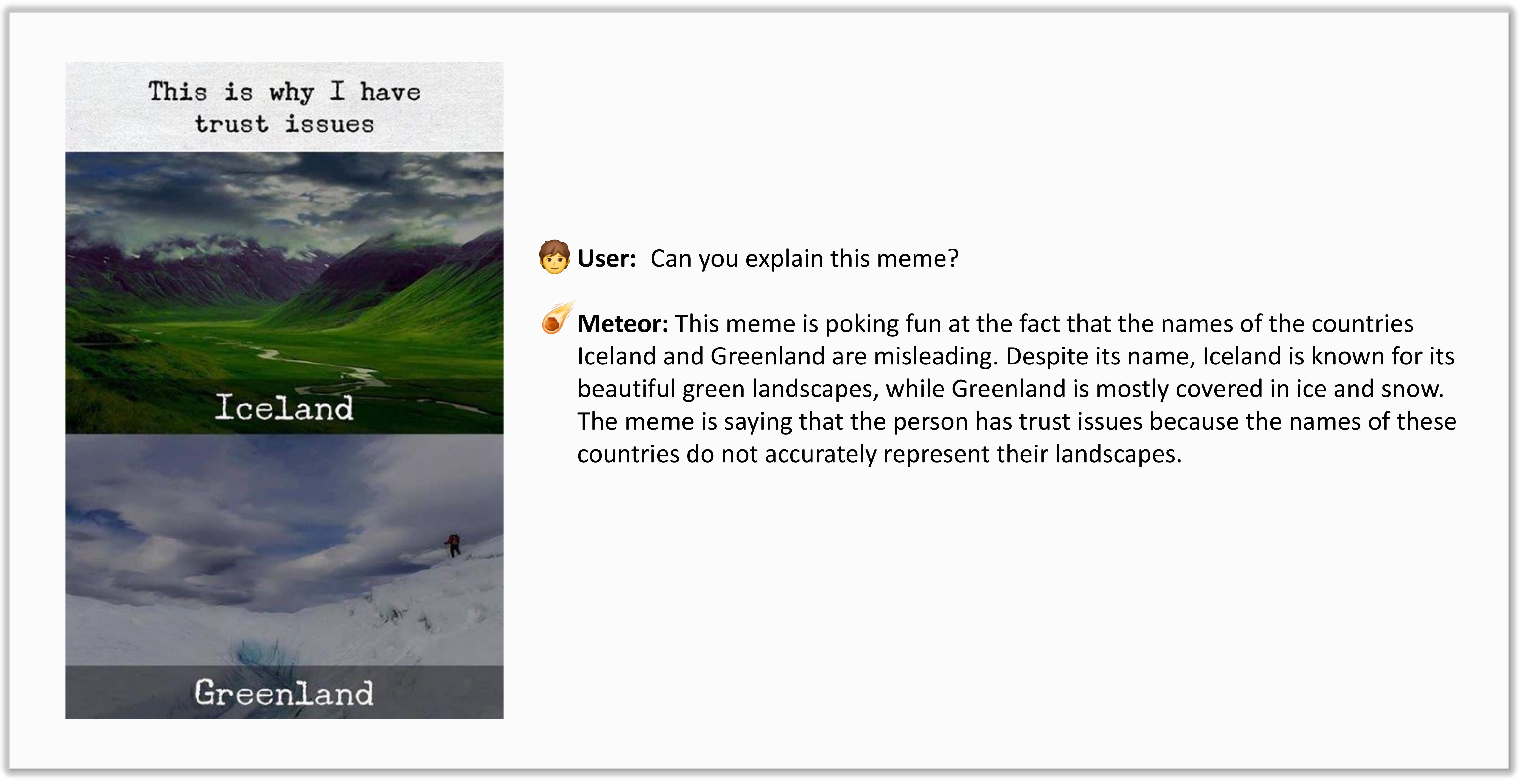}
\end{figure}

\begin{figure}[h!]
    \centering
    \includegraphics[width=\textwidth]{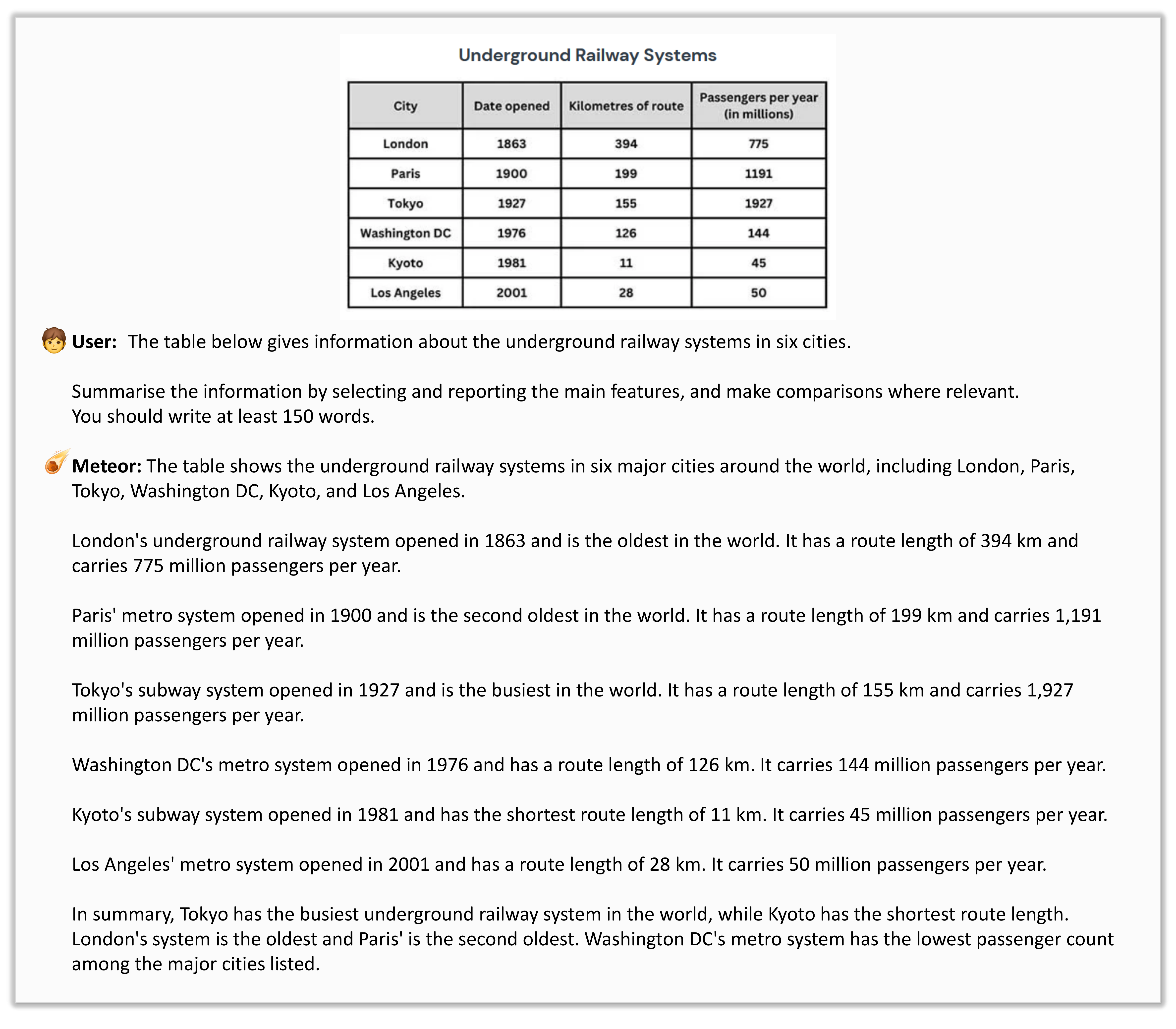}
\end{figure}

\begin{figure}[h!]
    \centering
    \includegraphics[width=\textwidth]{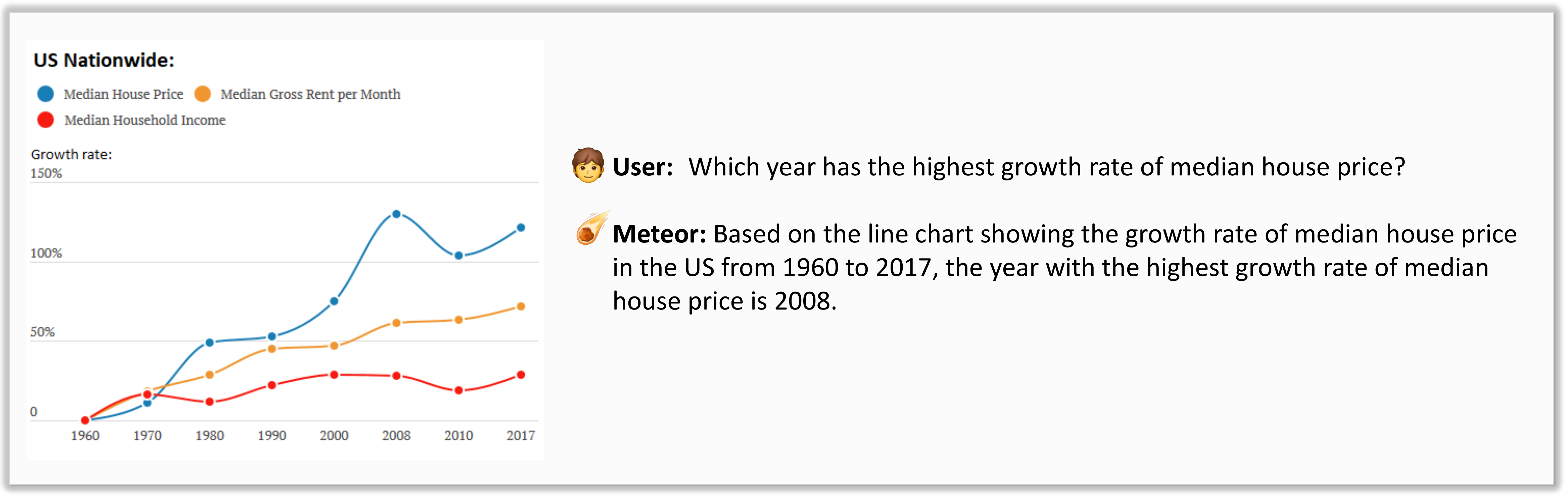}
\end{figure}

\begin{figure}[h!]
    \centering
    \includegraphics[width=\textwidth]{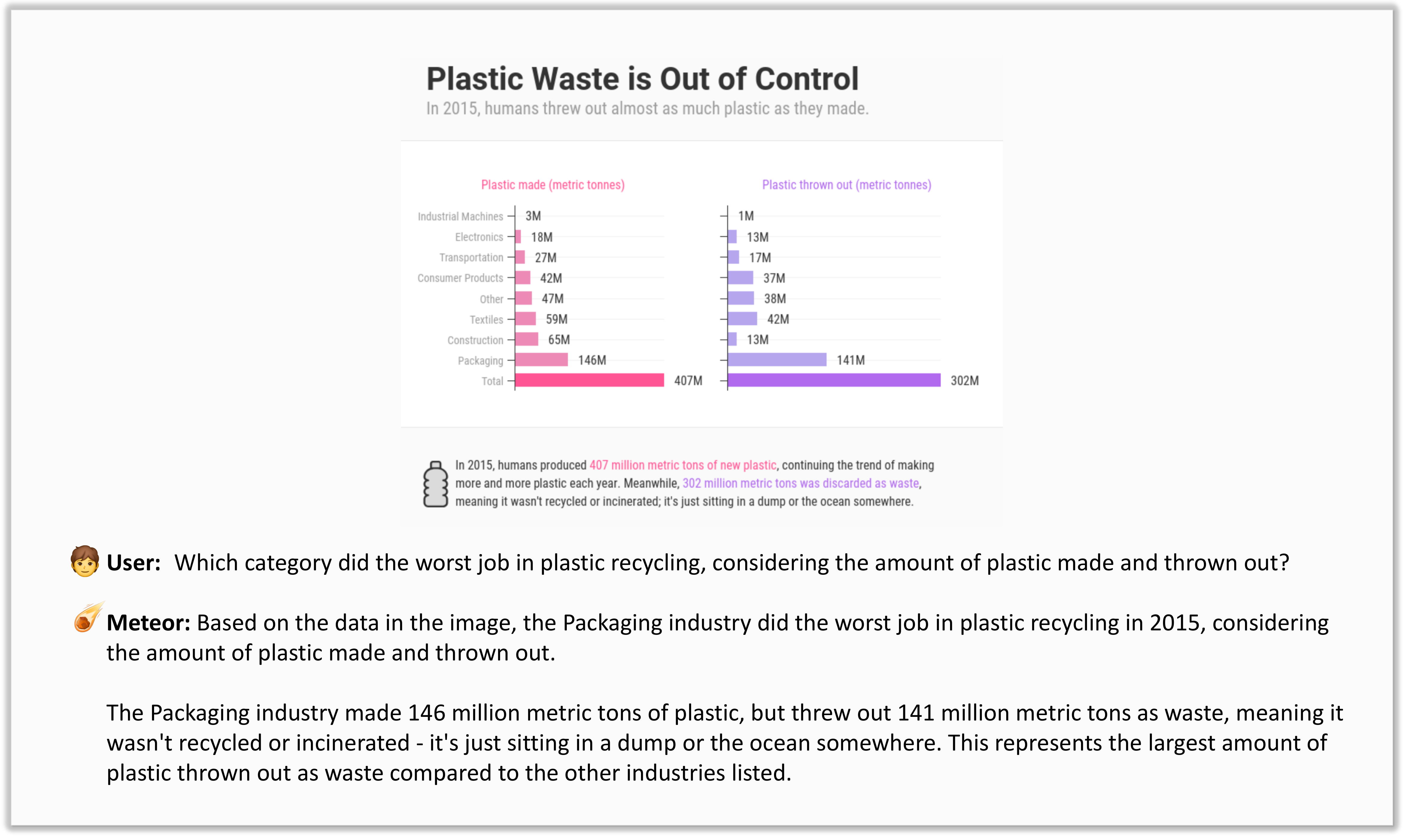}
\end{figure}

\clearpage

\begin{figure}[h!]
    \centering
    \includegraphics[width=\textwidth]{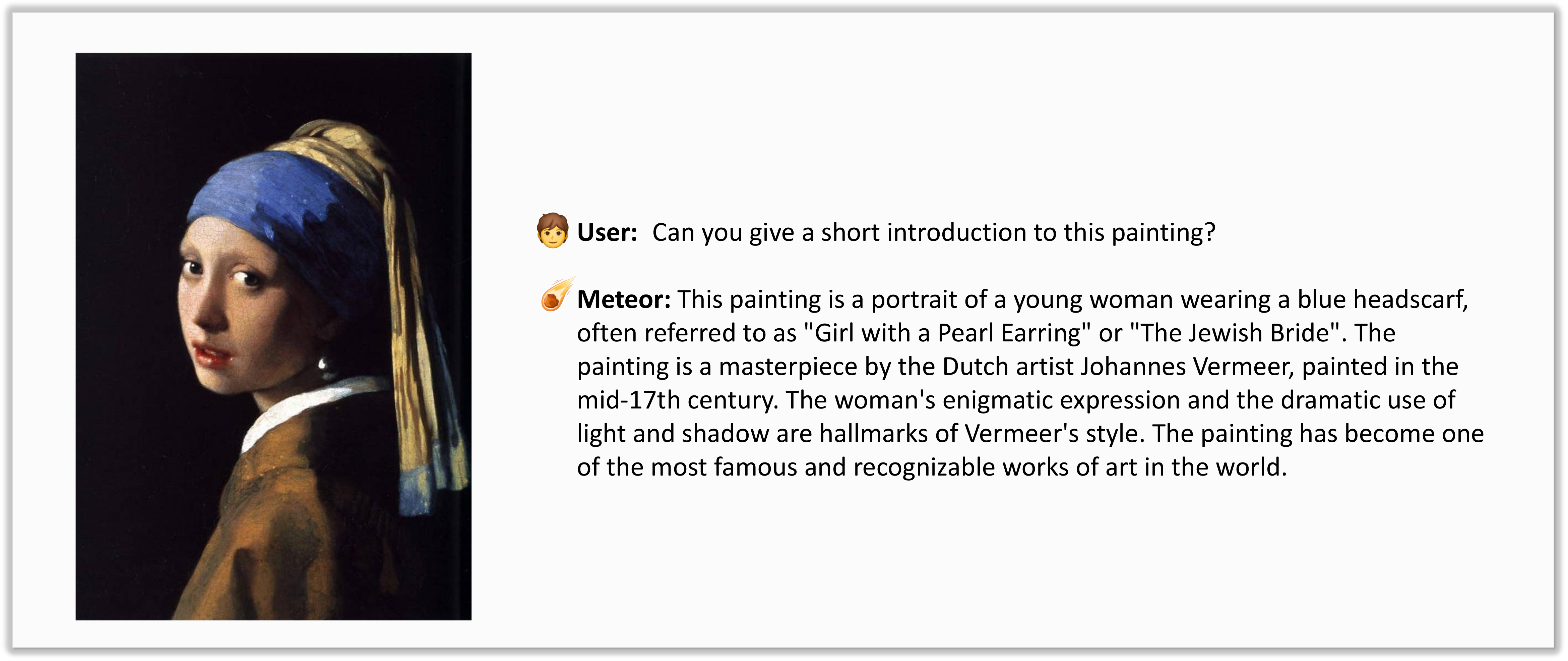}
\end{figure}

\begin{figure}[h!]
    \centering
    \includegraphics[width=\textwidth]{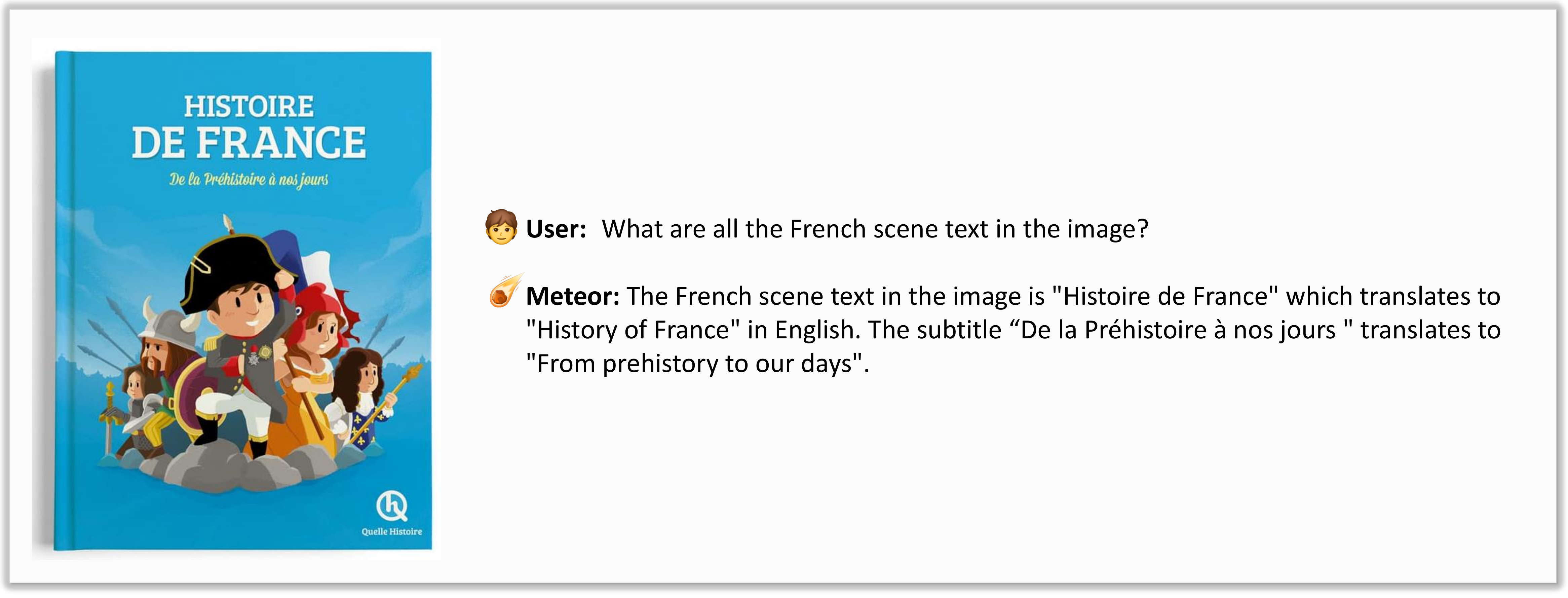}
\end{figure}

\clearpage
\section{Further Ablation Studies}
\label{sec:appD}
\begin{table}[h!]
\centering
\resizebox{\linewidth}{!}{
\renewcommand{\tabcolsep}{2mm}
\begin{tabular}{lcccccc}
\toprule
Methods         & AI2D & ChartQA & MathVista & MM-Vet & LLaVA$^{\text{W}}$ & MMStar \\ 
\midrule
Meteor          & 77.9          & 74.9             & 53.4               & 57.3            & 87.1            & 52.8            \\ 
Meteor-LLaVA-HR~\cite{llavahr} & 80.8      & 77.9             & 57.4               & 59.5            & 90.2            & 54.0            \\ 
\bottomrule
\end{tabular}
}
\vspace{2mm}
\caption{Performance comparison between \meteor Meteor and \meteor Meteor-LLaVA-HR across multiple tasks.}
\end{table}
\begin{table}[h!]
\resizebox{\linewidth}{!}{
\renewcommand{\tabcolsep}{3mm}
\centering
\begin{tabular}{lccccccc}
\toprule
Meteor-Mamba Size & AI2D & ChartQA & MathVista & MM-Vet & LLaVAW & MMStar \\
\midrule
130M  & 77.9 & 74.9 & 53.4 & 57.3 & 87.1 & 52.8 \\
790M  & 78.7 & 75.5 & 54.9 & 57.8 & 88.0 & 53.0 \\
1.4B  & 79.6 & 76.2 & 56.2 & 58.8 & 89.8 & 53.6 \\
\bottomrule
\end{tabular}
}
\vspace{2mm}
\caption{Performance of Meteor-Mamba models with different sizes across various tasks.}
\end{table}
\begin{table}[h!]
\centering
\resizebox{\linewidth}{!}{
\renewcommand{\tabcolsep}{5mm}
\begin{tabular}{lccccc}
\toprule
       & Qwen-VL & LLaVA1.5 & CoLLaVO & MoAI  & Meteor \\ 
\midrule
Time   & 16 toks/s & 22 toks/s & 21 toks/s & 20 toks/s & 22 toks/s \\
\bottomrule
\end{tabular}
}
\vspace{2mm}
\caption{Token processing speed comparison between models.}
\end{table}
\begin{table}[h!]
\centering
\resizebox{\linewidth}{!}{
\renewcommand{\tabcolsep}{3mm}
\begin{tabular}{lccccccc}
\toprule
LLVMs       & VQAv2 & GQA  & SQA-IMG & TextVQA & POPE  & MMB   & MM-Vet \\
\midrule
Cobra~\cite{zhao2024cobra}       & 76.9  & 59.9 & -       & 57.9    & 88.2  & -     & -      \\
VL-Mamba~\cite{qiao2024vlmamba}    & 76.6  & 56.2 & 65.4    & 48.9    & 84.4  & 57.0  & 32.6   \\
RoboMamba~\cite{liu2024robomamba}   & 79.1  & 64.4 & -       & -       & 86.9  & 65.7  & 29.7   \\
ML-Mamba~\cite{huang2024mlmamba}    & 75.3  & 60.7 & -       & 52.2    & 88.3  & -     & -      \\
\midrule
\rowcolor{Green}
Meteor      & \textbf{82.5}  & \textbf{64.7} & \textbf{88.3} & \textbf{67.5}    & \textbf{88.7}  & \textbf{82.9}  & \textbf{57.3}   \\
\bottomrule
\end{tabular}
}
\vspace{2mm}
\caption{Performance comparison of LLVMs across various tasks.}
\end{table}

\begin{table}[h!]
\centering
\resizebox{\linewidth}{!}{
\renewcommand{\tabcolsep}{2mm}
\begin{tabular}{lcccc}
\toprule
LLVMs                        & Conversation & Detail Description & Complex Reasoning & Avg  \\
\midrule
CoLLaVO                      & 51.1         & 73.8               & 77.1              & 69.5 \\
MoAI                         & 48.5         & 76.0               & 80.6              & 71.9 \\
\midrule
\rowcolor{Green}
Meteor w.o. Meteor-Mamba      & 67.4         & 72.9               & 75.2              & 73.7 \\
\rowcolor{Green}
Meteor                       & \textbf{80.3}         & \textbf{87.2}               & \textbf{91.2}              & \textbf{87.1} \\
\bottomrule
\end{tabular}
}
\vspace{2mm}
\caption{Performance comparison of LLVMs across Conversation, Detail Description, Complex Reasoning, and Average in LLaVA$^{\text{W}}$.}
\end{table}

\clearpage
\newpage
\section*{NeurIPS Paper Checklist}
\begin{enumerate}

\item {\bf Claims}
    \item[] Question: Do the main claims made in the abstract and introduction accurately reflect the paper's contributions and scope?
    \item[] Answer: \answerYes{} 
    \item[] Justification: Abstract and Introduction section
    \item[] Guidelines:
    \begin{itemize}
        \item The answer NA means that the abstract and introduction do not include the claims made in the paper.
        \item The abstract and/or introduction should clearly state the claims made, including the contributions made in the paper and important assumptions and limitations. A No or NA answer to this question will not be perceived well by the reviewers. 
        \item The claims made should match theoretical and experimental results, and reflect how much the results can be expected to generalize to other settings. 
        \item It is fine to include aspirational goals as motivation as long as it is clear that these goals are not attained by the paper. 
    \end{itemize}

\item {\bf Limitations}
    \item[] Question: Does the paper discuss the limitations of the work performed by the authors?
    \item[] Answer: \answerYes{} 
    \item[] Justification: Discussion and Limitation section
    \item[] Guidelines:
    \begin{itemize}
        \item The answer NA means that the paper has no limitation while the answer No means that the paper has limitations, but those are not discussed in the paper. 
        \item The authors are encouraged to create a separate "Limitations" section in their paper.
        \item The paper should point out any strong assumptions and how robust the results are to violations of these assumptions (e.g., independence assumptions, noiseless settings, model well-specification, asymptotic approximations only holding locally). The authors should reflect on how these assumptions might be violated in practice and what the implications would be.
        \item The authors should reflect on the scope of the claims made, e.g., if the approach was only tested on a few datasets or with a few runs. In general, empirical results often depend on implicit assumptions, which should be articulated.
        \item The authors should reflect on the factors that influence the performance of the approach. For example, a facial recognition algorithm may perform poorly when image resolution is low or images are taken in low lighting. Or a speech-to-text system might not be used reliably to provide closed captions for online lectures because it fails to handle technical jargon.
        \item The authors should discuss the computational efficiency of the proposed algorithms and how they scale with dataset size.
        \item If applicable, the authors should discuss possible limitations of their approach to address problems of privacy and fairness.
        \item While the authors might fear that complete honesty about limitations might be used by reviewers as grounds for rejection, a worse outcome might be that reviewers discover limitations that aren't acknowledged in the paper. The authors should use their best judgment and recognize that individual actions in favor of transparency play an important role in developing norms that preserve the integrity of the community. Reviewers will be specifically instructed to not penalize honesty concerning limitations.
    \end{itemize}

\item {\bf Theory Assumptions and Proofs}
    \item[] Question: For each theoretical result, does the paper provide the full set of assumptions and a complete (and correct) proof?
    \item[] Answer: \answerNA{} 
    \item[] Justification: None of theoretical assumptions
    \item[] Guidelines:
    \begin{itemize}
        \item The answer NA means that the paper does not include theoretical results. 
        \item All the theorems, formulas, and proofs in the paper should be numbered and cross-referenced.
        \item All assumptions should be clearly stated or referenced in the statement of any theorems.
        \item The proofs can either appear in the main paper or the supplemental material, but if they appear in the supplemental material, the authors are encouraged to provide a short proof sketch to provide intuition. 
        \item Inversely, any informal proof provided in the core of the paper should be complemented by formal proofs provided in appendix or supplemental material.
        \item Theorems and Lemmas that the proof relies upon should be properly referenced. 
    \end{itemize}

    \item {\bf Experimental Result Reproducibility}
    \item[] Question: Does the paper fully disclose all the information needed to reproduce the main experimental results of the paper to the extent that it affects the main claims and/or conclusions of the paper (regardless of whether the code and data are provided or not)?
    \item[] Answer: \answerYes{} 
    \item[] Justification: Meteor and Experiment section
    \item[] Guidelines:
    \begin{itemize}
        \item The answer NA means that the paper does not include experiments.
        \item If the paper includes experiments, a No answer to this question will not be perceived well by the reviewers: Making the paper reproducible is important, regardless of whether the code and data are provided or not.
        \item If the contribution is a dataset and/or model, the authors should describe the steps taken to make their results reproducible or verifiable. 
        \item Depending on the contribution, reproducibility can be accomplished in various ways. For example, if the contribution is a novel architecture, describing the architecture fully might suffice, or if the contribution is a specific model and empirical evaluation, it may be necessary to either make it possible for others to replicate the model with the same dataset, or provide access to the model. In general. releasing code and data is often one good way to accomplish this, but reproducibility can also be provided via detailed instructions for how to replicate the results, access to a hosted model (e.g., in the case of a large language model), releasing of a model checkpoint, or other means that are appropriate to the research performed.
        \item While NeurIPS does not require releasing code, the conference does require all submissions to provide some reasonable avenue for reproducibility, which may depend on the nature of the contribution. For example
        \begin{enumerate}
            \item If the contribution is primarily a new algorithm, the paper should make it clear how to reproduce that algorithm.
            \item If the contribution is primarily a new model architecture, the paper should describe the architecture clearly and fully.
            \item If the contribution is a new model (e.g., a large language model), then there should either be a way to access this model for reproducing the results or a way to reproduce the model (e.g., with an open-source dataset or instructions for how to construct the dataset).
            \item We recognize that reproducibility may be tricky in some cases, in which case authors are welcome to describe the particular way they provide for reproducibility. In the case of closed-source models, it may be that access to the model is limited in some way (e.g., to registered users), but it should be possible for other researchers to have some path to reproducing or verifying the results.
        \end{enumerate}
    \end{itemize}

\item {\bf Open access to data and code}
    \item[] Question: Does the paper provide open access to the data and code, with sufficient instructions to faithfully reproduce the main experimental results, as described in supplemental material?
    \item[] Answer: \answerYes{} 
    \item[] Justification: You can see the link in Abstract section.
    \item[] Guidelines:
    \begin{itemize}
        \item The answer NA means that paper does not include experiments requiring code.
        \item Please see the NeurIPS code and data submission guidelines (\url{https://nips.cc/public/guides/CodeSubmissionPolicy}) for more details.
        \item While we encourage the release of code and data, we understand that this might not be possible, so “No” is an acceptable answer. Papers cannot be rejected simply for not including code, unless this is central to the contribution (e.g., for a new open-source benchmark).
        \item The instructions should contain the exact command and environment needed to run to reproduce the results. See the NeurIPS code and data submission guidelines (\url{https://nips.cc/public/guides/CodeSubmissionPolicy}) for more details.
        \item The authors should provide instructions on data access and preparation, including how to access the raw data, preprocessed data, intermediate data, and generated data, etc.
        \item The authors should provide scripts to reproduce all experimental results for the new proposed method and baselines. If only a subset of experiments are reproducible, they should state which ones are omitted from the script and why.
        \item At submission time, to preserve anonymity, the authors should release anonymized versions (if applicable).
        \item Providing as much information as possible in supplemental material (appended to the paper) is recommended, but including URLs to data and code is permitted.
    \end{itemize}

\item {\bf Experimental Setting/Details}
    \item[] Question: Does the paper specify all the training and test details (e.g., data splits, hyperparameters, how they were chosen, type of optimizer, etc.) necessary to understand the results?
    \item[] Answer: \answerYes{} 
    \item[] Justification: Experiment section
    \item[] Guidelines:
    \begin{itemize}
        \item The answer NA means that the paper does not include experiments.
        \item The experimental setting should be presented in the core of the paper to a level of detail that is necessary to appreciate the results and make sense of them.
        \item The full details can be provided either with the code, in appendix, or as supplemental material.
    \end{itemize}

\item {\bf Experiment Statistical Significance}
    \item[] Question: Does the paper report error bars suitably and correctly defined or other appropriate information about the statistical significance of the experiments?
    \item[] Answer: \answerNo{} 
    \item[] Justification: Training Meteor with only one epoch under 4-bit quantized-extreme condition proves its effectiveness beyond checking error bar.
    \item[] Guidelines:
    \begin{itemize}
        \item The answer NA means that the paper does not include experiments.
        \item The authors should answer "Yes" if the results are accompanied by error bars, confidence intervals, or statistical significance tests, at least for the experiments that support the main claims of the paper.
        \item The factors of variability that the error bars are capturing should be clearly stated (for example, train/test split, initialization, random drawing of some parameter, or overall run with given experimental conditions).
        \item The method for calculating the error bars should be explained (closed form formula, call to a library function, bootstrap, etc.)
        \item The assumptions made should be given (e.g., Normally distributed errors).
        \item It should be clear whether the error bar is the standard deviation or the standard error of the mean.
        \item It is OK to report 1-sigma error bars, but one should state it. The authors should preferably report a 2-sigma error bar than state that they have a 96\% CI, if the hypothesis of Normality of errors is not verified.
        \item For asymmetric distributions, the authors should be careful not to show in tables or figures symmetric error bars that would yield results that are out of range (e.g. negative error rates).
        \item If error bars are reported in tables or plots, The authors should explain in the text how they were calculated and reference the corresponding figures or tables in the text.
    \end{itemize}

\item {\bf Experiments Compute Resources}
    \item[] Question: For each experiment, does the paper provide sufficient information on the computer resources (type of compute workers, memory, time of execution) needed to reproduce the experiments?
    \item[] Answer: \answerNo{} 
    \item[] Justification: Performances of LLVMs are not dependent on trivial hyper-parameter but dependent with propagation strategy, model architectures, and dataset. Many researcher for LLVMs have known that LLVMs can be instruction-tuned, at least, with multiple of GPUs such as NVIDIA RTX A6000 and A100.
    \item[] Guidelines:
    \begin{itemize}
        \item The answer NA means that the paper does not include experiments.
        \item The paper should indicate the type of compute workers CPU or GPU, internal cluster, or cloud provider, including relevant memory and storage.
        \item The paper should provide the amount of compute required for each of the individual experimental runs as well as estimate the total compute. 
        \item The paper should disclose whether the full research project required more compute than the experiments reported in the paper (e.g., preliminary or failed experiments that didn't make it into the paper). 
    \end{itemize}
    
\item {\bf Code Of Ethics}
    \item[] Question: Does the research conducted in the paper conform, in every respect, with the NeurIPS Code of Ethics \url{https://neurips.cc/public/EthicsGuidelines}?
    \item[] Answer: \answerYes{} 
    \item[] Justification: We used publicly common visual instruction tuning dataset.
    \item[] Guidelines:
    \begin{itemize}
        \item The answer NA means that the authors have not reviewed the NeurIPS Code of Ethics.
        \item If the authors answer No, they should explain the special circumstances that require a deviation from the Code of Ethics.
        \item The authors should make sure to preserve anonymity (e.g., if there is a special consideration due to laws or regulations in their jurisdiction).
    \end{itemize}

\item {\bf Broader Impacts}
    \item[] Question: Does the paper discuss both potential positive societal impacts and negative societal impacts of the work performed?
    \item[] Answer: \answerYes{} 
    \item[] Justification: Discussion and Limitation section
    \item[] Guidelines:
    \begin{itemize}
        \item The answer NA means that there is no societal impact of the work performed.
        \item If the authors answer NA or No, they should explain why their work has no societal impact or why the paper does not address societal impact.
        \item Examples of negative societal impacts include potential malicious or unintended uses (e.g., disinformation, generating fake profiles, surveillance), fairness considerations (e.g., deployment of technologies that could make decisions that unfairly impact specific groups), privacy considerations, and security considerations.
        \item The conference expects that many papers will be foundational research and not tied to particular applications, let alone deployments. However, if there is a direct path to any negative applications, the authors should point it out. For example, it is legitimate to point out that an improvement in the quality of generative models could be used to generate deepfakes for disinformation. On the other hand, it is not needed to point out that a generic algorithm for optimizing neural networks could enable people to train models that generate Deepfakes faster.
        \item The authors should consider possible harms that could arise when the technology is being used as intended and functioning correctly, harms that could arise when the technology is being used as intended but gives incorrect results, and harms following from (intentional or unintentional) misuse of the technology.
        \item If there are negative societal impacts, the authors could also discuss possible mitigation strategies (e.g., gated release of models, providing defenses in addition to attacks, mechanisms for monitoring misuse, mechanisms to monitor how a system learns from feedback over time, improving the efficiency and accessibility of ML).
    \end{itemize}
    
\item {\bf Safeguards}
    \item[] Question: Does the paper describe safeguards that have been put in place for responsible release of data or models that have a high risk for misuse (e.g., pretrained language models, image generators, or scraped datasets)?
    \item[] Answer: \answerNA{} 
    \item[] Justification: We used publicly common visual instruction tuning dataset and pre-trained visual foundation models and large language models.
    \item[] Guidelines:
    \begin{itemize}
        \item The answer NA means that the paper poses no such risks.
        \item Released models that have a high risk for misuse or dual-use should be released with necessary safeguards to allow for controlled use of the model, for example by requiring that users adhere to usage guidelines or restrictions to access the model or implementing safety filters. 
        \item Datasets that have been scraped from the Internet could pose safety risks. The authors should describe how they avoided releasing unsafe images.
        \item We recognize that providing effective safeguards is challenging, and many papers do not require this, but we encourage authors to take this into account and make a best faith effort.
    \end{itemize}

\item {\bf Licenses for existing assets}
    \item[] Question: Are the creators or original owners of assets (e.g., code, data, models), used in the paper, properly credited and are the license and terms of use explicitly mentioned and properly respected?
    \item[] Answer: \answerYes{} 
    \item[] Justification: We cited models and datasets we deal with in this paper.
    \item[] Guidelines:
    \begin{itemize}
        \item The answer NA means that the paper does not use existing assets.
        \item The authors should cite the original paper that produced the code package or dataset.
        \item The authors should state which version of the asset is used and, if possible, include a URL.
        \item The name of the license (e.g., CC-BY 4.0) should be included for each asset.
        \item For scraped data from a particular source (e.g., website), the copyright and terms of service of that source should be provided.
        \item If assets are released, the license, copyright information, and terms of use in the package should be provided. For popular datasets, \url{paperswithcode.com/datasets} has curated licenses for some datasets. Their licensing guide can help determine the license of a dataset.
        \item For existing datasets that are re-packaged, both the original license and the license of the derived asset (if it has changed) should be provided.
        \item If this information is not available online, the authors are encouraged to reach out to the asset's creators.
    \end{itemize}

\item {\bf New Assets}
    \item[] Question: Are new assets introduced in the paper well documented and is the documentation provided alongside the assets?
    \item[] Answer: \answerYes{} 
    \item[] Justification: Meteor and Experiment section
    \item[] Guidelines:
    \begin{itemize}
        \item The answer NA means that the paper does not release new assets.
        \item Researchers should communicate the details of the dataset/code/model as part of their submissions via structured templates. This includes details about training, license, limitations, etc. 
        \item The paper should discuss whether and how consent was obtained from people whose asset is used.
        \item At submission time, remember to anonymize your assets (if applicable). You can either create an anonymized URL or include an anonymized zip file.
    \end{itemize}

\item {\bf Crowdsourcing and Research with Human Subjects}
    \item[] Question: For crowdsourcing experiments and research with human subjects, does the paper include the full text of instructions given to participants and screenshots, if applicable, as well as details about compensation (if any)? 
    \item[] Answer: \answerNA{} 
    \item[] Justification: None of any crowdsourcing and research with human subjects
    \item[] Guidelines:
    \begin{itemize}
        \item The answer NA means that the paper does not involve crowdsourcing nor research with human subjects.
        \item Including this information in the supplemental material is fine, but if the main contribution of the paper involves human subjects, then as much detail as possible should be included in the main paper. 
        \item According to the NeurIPS Code of Ethics, workers involved in data collection, curation, or other labor should be paid at least the minimum wage in the country of the data collector. 
    \end{itemize}

\item {\bf Institutional Review Board (IRB) Approvals or Equivalent for Research with Human Subjects}
    \item[] Question: Does the paper describe potential risks incurred by study participants, whether such risks were disclosed to the subjects, and whether Institutional Review Board (IRB) approvals (or an equivalent approval/review based on the requirements of your country or institution) were obtained?
    \item[] Answer: \answerNA{} 
    \item[] Justification: None of any crowdsourcing and research with human subjects
    \item[] Guidelines:
    \begin{itemize}
        \item The answer NA means that the paper does not involve crowdsourcing nor research with human subjects.
        \item Depending on the country in which research is conducted, IRB approval (or equivalent) may be required for any human subjects research. If you obtained IRB approval, you should clearly state this in the paper. 
        \item We recognize that the procedures for this may vary significantly between institutions and locations, and we expect authors to adhere to the NeurIPS Code of Ethics and the guidelines for their institution. 
        \item For initial submissions, do not include any information that would break anonymity (if applicable), such as the institution conducting the review.
    \end{itemize}

\end{enumerate}

\end{document}